\documentclass[sigconf]{acmart}

\usepackage{subcaption}

\usepackage{arydshln}
\usepackage{tikz}
\usepackage{pifont}
\usepackage{multirow}
\usepackage{amsmath,amsfonts}
\usepackage{booktabs}
\usepackage[most]{tcolorbox}
\newtcolorbox{promptbox}{
colback=gray!5,  
colframe=black!75, 
left=1em, 
right=1em, 
top=1em, 
bottom=1em, 
sharp corners, 
boxrule=1pt 
}
\usepackage{algorithm}
\usepackage{algorithmic}
\usepackage{caption}
\AtBeginDocument{%
  }

\setcopyright{acmlicensed}
\copyrightyear{2018}
\acmYear{2018}
\acmDOI{XXXXXXX.XXXXXXX}

\acmConference[Conference acronym 'XX]{Make sure to enter the correct
  conference title from your rights confirmation emai}{June 03--05,
  2018}{Woodstock, NY}
\acmISBN{978-1-4503-XXXX-X/18/06}

\begin{document}

\title{Assessing the Impact of Conspiracy Theories Using Large Language Models}


\author{Bohan Jiang}
\affiliation{%
  \institution{Arizona State University}
  \city{Tempe}
  \state{Arizona}
  \country{USA}}
\email{bjiang14@asu.edu}

\author{Dawei Li}
\affiliation{%
  \institution{Arizona State University}
  \city{Tempe}
  \state{Arizona}
  \country{USA}}
\email{daweili5@asu.edu}

\author{Zhen Tan}
\affiliation{%
  \institution{Arizona State University}
  \city{Tempe}
  \state{Arizona}
  \country{USA}}
\email{ztan36@asu.edu}

\author{Xinyi Zhou}
\affiliation{%
  \institution{University of Washington}
  \city{Seattle}
  \state{Washington}
  \country{USA}}
\email{xzhou59@uw.edu}

\author{Ashwin Rao}
\affiliation{%
  \institution{USC Information Sciences Institute}
  \city{Marina Del Rey}
  \state{California}
  \country{USA}}
\email{mohanrao@usc.edu}

\author{Kristina Lerman}
\affiliation{%
  \institution{USC Information Sciences Institute}
  \city{Marina Del Rey}
  \state{California}
  \country{USA}}
\email{lerman@usc.edu}

\author{H. Russell Bernard}
\affiliation{%
  \institution{ASU Institute for Social Science Research}
  \city{Tempe}
  \state{Arizona}  
  \country{USA}}
\email{asuruss@asu.edu}

\author{Huan Liu}
\affiliation{%
  \institution{Arizona State University}
  \city{Tempe}
  \state{Arizona}  
  \country{USA}}
\email{huanliu@asu.edu}

\renewcommand{\shortauthors}{Jiang et al.}

\begin{abstract}
Measuring the relative impact of CTs is important for prioritizing responses and allocating resources effectively, especially during crises. However, assessing the actual impact of CTs on the public poses unique challenges. It requires not only the collection of CT-specific knowledge but also diverse information from social, psychological, and cultural dimensions. 
Recent advancements in large language models (LLMs) suggest their potential utility in this context, not only due to their extensive knowledge from large training corpora but also because they can be harnessed for complex reasoning. In this work, we \textbf{\textit{develop}} datasets of popular CTs with human-annotated impacts. Borrowing insights from human impact assessment processes, we then \textbf{\textit{design}} tailored strategies to leverage LLMs for performing human-like CT impact assessments. Through rigorous experiments, we \textbf{\textit{discover}} that \ding{182}~an impact assessment mode using multi-step reasoning to analyze more CT-related evidence critically produces accurate results; and \ding{183}~most LLMs demonstrate strong bias, such as assigning higher impacts to CTs presented earlier in the prompt, while generating less accurate impact assessments for emotionally charged and verbose CTs.
\end{abstract}




\maketitle

\section{Introduction}
\begin{figure*}[t]
\centering
\includegraphics[width=1\textwidth]{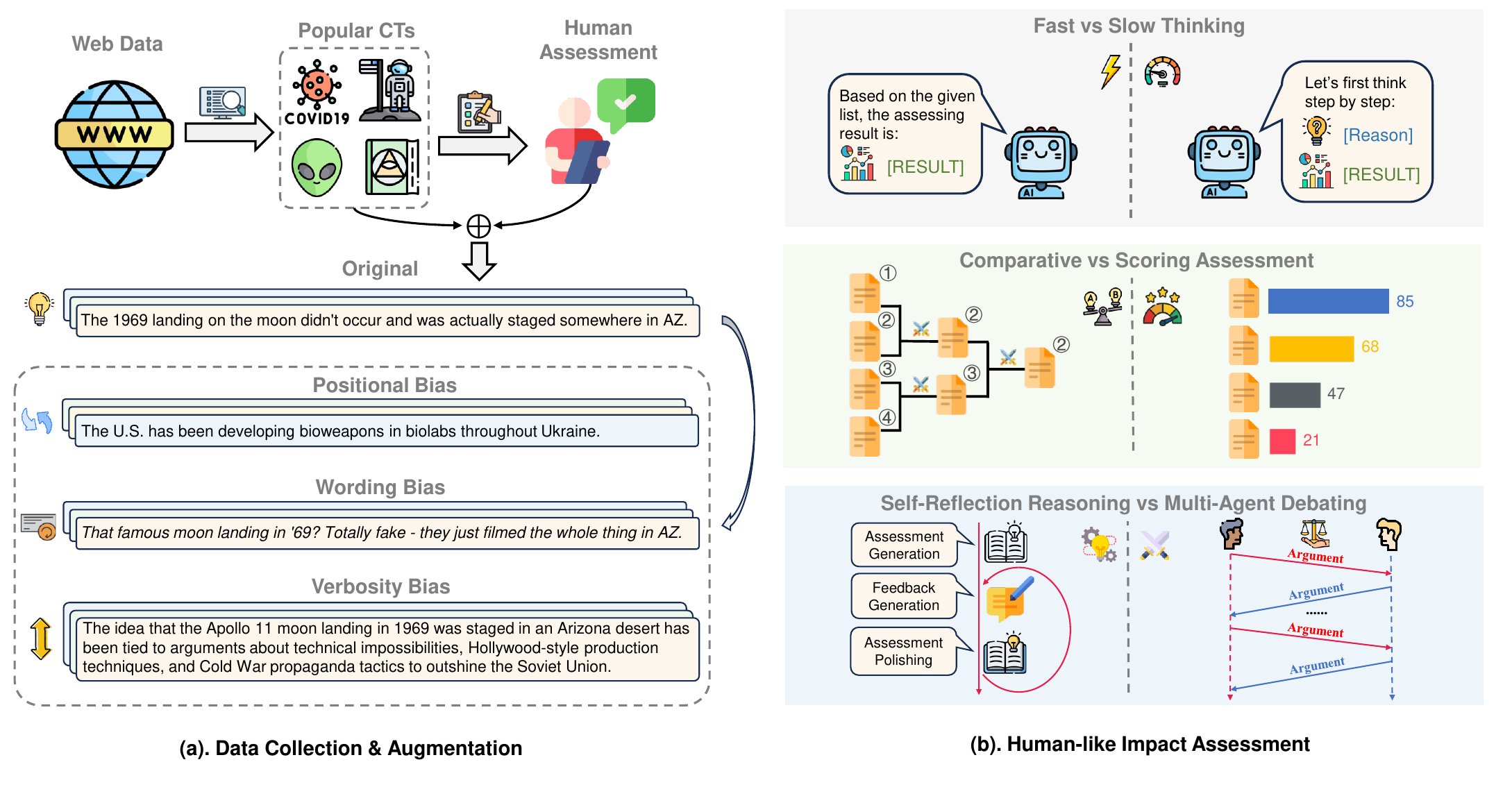} 
\caption{Research pipeline for CT impact assessment using LLMs. The pipeline consists of two main stages: (a) Data Collection and Augmentation, where a list of popular conspiracy theories with human-annotated impact assessments is expanded using position, wording, and verbosity perturbations to test robustness; and (b) Human-like Impact Assessment, leveraging tailored prompting strategies to simulate human reasoning processes, including fast versus slow thinking, comparative versus scoring assessments, and single-agent versus multi-agent debating.}
\label{fig:overview}
\end{figure*}
Conspiracy theories (CTs) are beliefs that social events and circumstances are secretly controlled by powerful groups~\cite{sunstein2009conspiracy}. Unlike general misinformation and disinformation~\cite{lazer2018science}, CTs stand out due to their profound real-world impacts~\cite{douglas2019understanding, cassam2019conspiracy}, such as fueling distrust in institutions, inciting tribalism, and provoking violence~\cite{freeman2022coronavirus, jolley2020pylons,gallacher2021online}. For example, one CT claimed that COVID-19 vaccines contained microchips designed by tech companies to track individuals' personal data~\cite{goodman2020coronavirus}, resulting in vaccine refusal and offline violence~\cite{pertwee2022epidemic,romer2020conspiracy}. Moreover, social media has further amplified the spread of CTs by attracting and connecting like-minded believers, providing fertile ground for CTs to proliferate~\cite{bavel2020using, jiang2021mechanisms}.

In response, researchers and tech companies have developed intelligent systems to detect and remove CT-related content from online spaces~\cite{shahsavari2020conspiracy,diab2024classifying, liaw2023younicon}. However, such content moderation efforts have raised significant concerns about freedom of speech and have led to unintended negative outcomes~\cite{innes2023platforming}. According to \textit{psychological reactance} theory~\cite{brehm1966theory}, restricting access to information can backfire by sparking curiosity about the prohibited content, potentially reinforcing CT communities~\cite{monti2023online}. Therefore, rather than adopting a blanket ``remove-all'' strategy, \textit{it is more effective to prioritize combating the most impactful CTs}. This is particularly important during emerging crises, where limited resources should be allocated to CTs with the greatest potential harm.

Assessing the impact of CTs is a challenging task. It typically requires human annotators with substantial background knowledge and access to extensive supporting evidence to evaluate CTs across various societal and psychological dimensions~\cite{douglas2017psychology}. Although researchers have created guidelines for impact assessment~\cite{esteves2012social, assessment1995guidelines}, they are labor-intensive and lack scalability. This raises a key question: \textit{can machines be leveraged for CT impact assessment?} Intuitively, large language models (LLMs) stand out as a promising tool due to their vast training datasets and extensive knowledge of social events. Recent advancements in LLMs, such as OpenAI's GPT series~\cite{achiam2023gpt, hurst2024gpt, openai2024o1preview} and Meta's LLaMA series~\cite{touvron2023llama, touvron2023llama2, dubey2024llama}, have demonstrated great proficiency in analyzing social problems and evaluating natural language processing (NLP) and machine learning (ML) tasks~\cite{kojima2022large,tong2024can}. For example, recent studies have explored the LLMs' potential to detect misinformation and disinformation in a zero-shot setting~\cite{chen2023can, jiang2024disinformation}, as well as their ability to simulate complex social interactions through role-playing human agents~\cite{zhou2023sotopia, park2023generative}. Other research has developed assessment and evaluation methods using LLMs, such as LLM-as-a-judge~\cite{li2024generation}, LLM-as-an-examiner~\cite{bai2024benchmarking}, and human-LLM collaborative evaluation~\cite{gao2024taxonomy}. Despite these advancements, using LLMs to assess the impact of CTs requires more complex reasoning, vast and diverse information, and a deep understanding of social dynamics and human cognition, an area that remains largely unexplored.

To bridge this gap, we aim to investigate the feasibility of \textbf{leveraging LLMs for automating impact assessment of conspiracy theories}. Specifically, this work evaluates the performance of LLMs in CT impact assessment by comparing their outputs to human-annotated results (gold labels). To do so, we first develop a CT dataset based on the 2023 YouGov survey~\cite{yougov2023conspiracy}, which collects public belief in a diverse range of CTs from a representative sample of 1,000 U.S. adults. Next, drawing inspiration from social science guidelines for impact assessment~\cite{esteves2012social}, we design tailored strategies that guide LLMs to simulate human-like CT impact assessment. Particularly, we purposely harness the LLM to mimic distinct human thinking processes (fast versus slow thinking), assessment paradigms (comparative versus absolute scoring assessment), and interactive behaviors (self-reflection reasoning versus multi-agent debating). We conduct experiments on eight LLMs, using small and large, open-source and proprietary LLMs. Our empirical findings reveal that a multi-step impact assessment pipeline, which incorporates iterative information extraction and critical reasoning using multi-agent debating, produces more accurate assessments. Last but not least, we rigorously examine the robustness of their CT impact assessment capabilities against various prompting biases, including position bias, wording bias, and verbosity bias. We observe that most LLMs tend to assign disproportionately higher impacts to CTs appearing in the front position in the prompt. Moreover, emotionally charged and verbose CTs with irrelevant information have negatively influenced LLMs' CT impact assessment results. In summary, the contributions of this paper are as follows:

\begin{itemize}
    \item \textbf{Introduction of a novel task and dataset:} We propose the task of CT impact assessment using LLMs and develop datasets for this purpose.
    \item \textbf{Human-like CT impact assessment:} We design tailored strategies based on impact assessment guidelines designed by social scientists. We use these strategies to harness LLMs to perform human-like CT impact assessment.
    \item \textbf{Comprehensive experiment and insights:} We conduct extensive experiments to evaluate the general effectiveness of LLMs in CT impact assessment and investigate their robustness against different prompting biases. We highlight key findings and insights based on our experimental results.
\end{itemize}

\section{Related Work}
\subsection{LLMs for Assessment}
Assessment and evaluation have brought consistent challenges in artificial intelligence (AI) and machine learning (ML)~\cite{papineni2002bleu,lin2004rouge}. 
While traditional assessment approaches heavily rely on reference annotation and human supervision, the recent advancement in LLMs has inspired methods that adopt LLMs for assessment and judgement~\cite{li2024generation,li2024dalk}.
LLM-based assessments have been widely adopted in various NLP tasks such as translation~\cite{kocmi2023large}, summarization~\cite{gao2023human}, open-ended generation~\cite{zheng2023judging} and alignment~\cite{wang2024bpo}, leveraging well-designed judgment pipelines, leading to a huge improvement in efficiency and scalability.
Besides, LLM-based assessment has also been employed in many real-world applications that require human-like planning and reasoning capabilities.
\cite{liu2023reviewergpt} first propose to utilize LLMs in paper review and assessment, introducing three key tasks to study the LLMs' capabilities in paper reviewing: error identification, checklist verification, and better paper choosing.
Based on their experiment, they conclude that LLMs have a promising use as reviewing assistants for paper quality assessment.
Following them, many other works explore various pipelines~\cite{liang2024can}, benchmarks~\cite{zhou2024llm} and challenges~\cite{ye2024we} for LLM-based paper reviewing.
Besides, code assessment and evaluation have long been critical challenges in computer science.
Recently, with the promising code understanding and generation performance of LLMs, some studies have started to leverage LLMs for automatic code assessment.
\cite{mcaleese2024llm} first propose to train ``critic'' LLMs that help humans more accurately and efficiently evaluate model-written code.
\cite{zhao2024codejudge} introduce CodeJudge-Eval (CJ-Eval), which is a novel benchmark designed to evaluate LLMs’ code comprehension abilities from the perspective of code judging and assessment rather than generation.
Additionally, LLM-based assessment is also used in other scenarios and applications, including medical decision-making~\cite{wang2024llm}, retrieval systems~\cite{li2024dalk} and content moderation~\cite{beigi2024lrq}.
In this work, we borrow insight from these works and adopt LLMs' judgment ability in CTs' impact assessment, a critical and fundamental task for social good.

\subsection{LLMs for Social Problems}
There is growing interest in addressing social problems with LLMs. One area that significantly benefits from it is fact-checking, where LLMs have been widely utilized for misinformation detection, attribution, and mitigation~\cite{jiang2024disinformation, chen2023can,beigi2024lrq}. Other researchers have explored using LLMs for online content analysis. \cite{lyu2023gpt} first propose to employ GPT-4V as a social media content analysis engine, performing tasks including sentiment analysis, hate speech detection, fake news classification, demographic inference, and political ideology detection. \cite{yu2024popalm} propose Popularity-Aligned Language Models (PopALM), training LLMs with popular and unpopular comments for trendy response prediction in social media. Besides, many works leverage agent-based LLMs for social media simulation. \cite{tornberg2023simulating} first propose to leverage LLM agents to help researchers study how different news feed algorithms shape the quality of online conversations. Moreover, studies use LLM agents to perform various social simulations~\cite{ziems2024can,huang2024social}. \cite{gao2023s} construct the S3 system (short for Social network Simulation System), observing the emergence of population-level phenomena, including the propagation of information, attitudes, and emotions. Recently, \cite{yang2024oasis} devised OASIS, a generalizable and scalable social media simulator, replicating various social phenomena, including information spreading, group polarization, and herd effects across X and Reddit platforms.

\section{Datasets Development}
Developing labeled datasets for CT impact assessment is the most important step for evaluating the capabilities of LLMs. In this section, we detailed the data collection and augmentation processes. As shown in Table~\ref{tab:dataset_statistics}, the datasets comprises a default set of human-annotated CTs and three augmented subsets designed to evaluate the robustness of LLMs against position, wording, and verbosity biases in the prompts. 
\begin{table}[ht]
\centering
\caption{Datasets Statistics}
\begin{tabular}{|l|c|c|c|}
\hline
\textbf{Dataset} & \textbf{\# of CTs} & \textbf{Avg. Length} & \textbf{Labeled} \\ \hline
Default ($D$) & 12 & 12.6 & \checkmark \\ \hline
Position Bias ($D_p$) & 144 (12*12) & 12.6 & \checkmark \\ \hline
Wording Bias ($D_w$) & 36 (12*3) & 14.8 & \checkmark \\ \hline
Verbosity Bias ($D_v$) & 24 (12*2) & 36.3 & \checkmark \\ \hline
\end{tabular}
\label{tab:dataset_statistics}
\end{table}
\subsection{Survey Data}
A dataset of human annotations (ground truth) on the impact of CTs is pivotal for evaluating LLM outputs. However, accurately assessing the impact of CTs across diverse social, economic, cultural, and historical contexts presents significant challenges. In practice, researchers use the \textit{volume of faithful CT believers} as a proxy for estimating broader societal impact~\cite{goertzel1994belief, enders2022relationship, romer2021patterns}. This approach is grounded in a clear rationale: the larger the population believing in a CT, the greater its potential for tangible real-world impacts~\cite{douglas2019understanding}.

In this study, we utilize data from a recent YouGov survey that collects public belief in a wide range of prominent CTs. According to Media Bias/Fact Checking~\cite{mediabias_yougov}, YouGov was rated as ``Very High'' in factual reporting and overall ``least biased'' in U.S. polling. The survey sampled 1,000 U.S. adults, with a margin of error of ±4\%. Participants were selected from YouGov's opt-in online panel using sample matching and were weighted based on key demographic factors such as gender, age, race, etc. This stratification ensures the sample is representative of the broader U.S. population to mitigate potential selection bias.

Respondents were asked to evaluate their beliefs in 12 CTs, rating each as definitely true, probably true, probably false, definitely false, or unsure. These CTs covered a broad spectrum of conspiracy narratives, including governments and powerful groups controlling global events, microchips in vaccines, and skepticism surrounding the 1969 moon landing. Although it is hard to reflect participants' deeper actual beliefs through survey questions, strong perceived beliefs also lead to psychological and behavioral changes~\cite{chen2020knowledge}. Thus, we rank the impact of CTs based on the percentage of respondents who rated them as ``definitely true'' or ``probably true''.

\subsection{Prompting Bias Datasets}
To evaluate the robustness of LLMs against various prompting biases in assessing the impact of CTs, we introduce controlled perturbations to augment the original human-annotated CT impact dataset. Let \( D = \{CT_1, CT_2, \ldots, CT_N\} \) represent the orginal dataset, where each \( CT_i \) has an associated original human-assessed impact ranking \(y_i\). We keep \(y_i\) unchanged for the augmented versions of \(CT_i\) because all data augmentation methods we used preserve the original semantic information of \(CT_i\).

\paragraph{\textbf{Position Bias Dataset (\( D_p \))}.} 
Position bias arises when the order in which options or sequences of information are presented affects LLM's predicted impact rankings. We generate the position bias dataset \( D_p \) by shuffling the order of CTs in the original dataset \(D\). For example:
\[
D_{p}^m = \{CT_9, CT_4, \ldots, CT_N, CT_2\},
\]
where \(m\) denotes a specific permutation. The final dataset for testing position bias is: \[D_p = \{D_{p}^1, D_{p}^2, \ldots, D_{p}^m\}. \]

\paragraph{\textbf{Wording Bias Dataset (\( D_w \))}.}
Wording bias occurs when differences in language style or phrasing influence the LLM's impact assessment. For each \( CT_i \), we create rephrased variations in three styles, including formal (\( CT_i^f \)), casual (\( CT_i^c \)), and neutral (\( CT_i^n \)). The original dataset is augmented with these variations:
\begin{align*}
D_{w}^f & = \{CT_1^f, CT_2^f, \ldots, CT_N^f\}, \\
D_{w}^c & = \{CT_1^c, CT_2^c, \ldots, CT_N^c\}, \\
D_{w}^n & = \{CT_1^n, CT_2^n, \ldots, CT_N^n\}.
\end{align*}
The complete wording bias testing dataset is:
\[
D_{w} = \{D_w^f, D_w^c, D_w^n\}.
\]

\paragraph{\textbf{Verbosity Bias Dataset (\( D_v \))}.} 
Verbosity bias occurs when unnecessarily verbose, complex, and longer input text skews the LLM's impact assessment of the CTs. For each \( CT_i \), we generate two variations: one with contextually relevant verbosity \( CT_i^{re} \) and another with contextually irrelevant verbosity \( CT_i^{ir} \). The augmented datasets for testing verbosity bias are:
\begin{align*}
D_{v}^{re} & = \{CT_1^{re}, CT_2^{re}, \ldots, CT_N^{re}\}, \\
D_{v}^{ir} & = \{CT_1^{ir}, CT_2^{ir}, \ldots, CT_N^{ir}\}.
\end{align*}
The final verbosity bias testing dataset combines these variations:
\[
D_{v} = \{D_v^{re}, D_v^{ir}\}.
\]

By creating the position bias dataset \( D_p \), wording bias dataset \( D_w \), and verbosity bias dataset \( D_v \), we can systematically probe into vulnerabilities in LLMs during CT impact assessment.

\section{Human-like Impact Assessment}
\begin{figure*}[h]\small
    \centering
    
    \begin{tcolorbox}[colback=yellow!10!white,colframe=gray!50!white,title={\textsc{Templates}:}]

    \textbf{Vanilla Ranking. }\\ \textit{Given the following [\textbf{list of CTs}], please rank them from most to least impactful. Please only output the final ranking.}
    \\
    \tikz{\draw[dashed] (0,0) -- (\linewidth,0);} 
    
    \textbf{Chain-of-Thought. }\\ \textit{Given the following [\textbf{list of CTs}], please rank them from most to least impactful, providing reasoning as intermediate steps followed by the final ranking.}
    \\
    \tikz{\draw[dashed] (0,0) -- (\linewidth,0);} 

    \textbf{Pairwise Comparison. }\\ \textit{Given two conspiracy theories at a time from the following [\textbf{list of CTs}], compare their relative impact and decide which one is more impactful.}
    \\
    \tikz{\draw[dashed] (0,0) -- (\linewidth,0);} 

    \textbf{Individual Scoring. }
    \\ \textit{For each conspiracy theory in the following [\textbf{list of CTs}], assign an individual impact score on a scale of 1 to 100, where 100 indicates maximum impact.}
    \tikz{\draw[dashed] (0,0) -- (\linewidth,0);} 

    \textbf{Multi-Agent Debating. }
    \\ \textit{<Base Solution Generation>: Given the following conspiracy theories [\textbf{list of CTs}], please rank them from most to least impactful.}
    \\ \textit{<Debating>: You are a debater. The debate topic is stated as follows: What is the correct impact ranking for the following conspiracy list: [\textbf{list of CTs}]. Another debater's solution is [\textbf{Another Debater's Solution}]. You disagree with this solution. Provide your solution and reasons.}
     \\ \textit{<Judgment>: You are a moderator. Two debaters will present their solutions and discuss their perspectives on the correct answer: [\textbf{list of CTs}].  Affirmative side arguing: [\textbf{Affirmative Side's Solution}]. Negative side arguing: [\textbf{Negative Side's Solution}]. You will evaluate both sides' solutions and determine if there is a clear preference for a solution candidate.}

    \tikz{\draw[dashed] (0,0) -- (\linewidth,0);} 
    
     \textbf{Self-Reflection Reasoning. }
    \\ \textit{<Assessment Generation>: Given the following conspiracy theories, please rank them from most to least impactful with the original index numbers preserved: [\textbf{list of CTs}].}
    \\ \textit{<Reflection Generation>: Given the following conspiracy theories and a given ranking list based on their impact, please produce a critique and reflection on it, and decide whether it can be used as the final solution. Conspiracy: [\textbf{list of CTs}], Original Solution: [\textbf{current assessment}].}
     \\ \textit{<Assessment Polishing>: Given the following conspiracy theories, please rank them from most to least impactful with the original index numbers preserved.
    There is already an original solution and reflection on this solution, please polish the given solution referring to the reflection. Conspiracy: [\textbf{list of CTs}], Original Solution: [\textbf{current assessment}], Reflection: [\textbf{generated reflection}].}
    \end{tcolorbox}
  
    \caption{Templates for human-like CT impact assessment.}
    \label{fig:prompt_temp}
\end{figure*}
As shown in Figure~\ref{fig:prompt_temp}, we design tailored strategies to harness LLMs for simulating human-like impact assessments. These strategies evaluate LLMs' CT impact assessment capability in three dimensions: (1) thinking processes (fast versus slow thinking), (2) impact assessment paradigms (comparative versus absolute scoring assessment), and (3) interactive behaviors (self-reflection reasoning versus multi-agent debating).

\subsection{Fast and Slow Thinking}
Inspired by psychological theories, human cognition can be divided into two systems \cite{daniel2017thinking}. The fast thinking system is characterized by rapid, instinctive, and heuristic-based inference. In contrast, the slow thinking system involves deliberate analysis and logical reasoning. To simulate these scenarios, we design the following prompting strategies:

\paragraph{\textbf{Vanilla Ranking (Fast Thinking)}}
The LLM is prompted to directly produce a ranking of \( \{CT_1, CT_2, \ldots, CT_n\} \) without generating any reasoning or explanation. The output is the final LLM-predicted CT impact ranking \( \{\pi_1, \pi_2, \ldots, \pi_n\} \) such that:
\[
\pi_i < \pi_j \implies \text{Impact}(CT_{i}) > \text{Impact}(CT_{j}).
\]

\paragraph{\textbf{Chain-of-Thought (Slow Thinking)}}
The LLM is prompted to generate intermediate reasoning steps before providing a final ranking. The output includes a sequence of intermediate reasoning steps \( R_k \) for \( k \in \{1, 2, \ldots, n\} \), followed by a final CT impact ranking \( \{\pi_1, \pi_2, \ldots, \pi_n\} \). This encourages the model to use logical reasoning, reflecting deliberate decision-making processes.

\subsection{Comparative and Scoring Assessment}
We explore two paradigms for assessing CT impact: \textit{comparison} (Pairwise Comparison) and \textit{scoring} (Individual Scoring).

\paragraph{\textbf{Pairwise Comparison.}}
The LLM compares two CTs, \( CT_i \) and \( CT_j \), and selects the more impactful one. This process requires \(\binom{n}{2}\) comparisons for \( n \) CTs to construct the global CT impact ranking:
\[
\text{Compare}(CT_i, CT_j) = 
\begin{cases} 
CT_i & \text{if } \text{Impact}(CT_i) > \text{Impact}(CT_j), \\
CT_j & \text{otherwise}.
\end{cases}
\]
The results are aggregated to derive the global CT impact ranking.

\paragraph{\textbf{Individual Scoring.}}
The LLM assigns an individual impact score \( s_i \) to each \( CT_i \), where:
\[
s_i = \text{ImpactScore}(CT_i), \quad i \in \{1, 2, \ldots, n\}.
\]
The final CT impact ranking is derived by sorting the CTs based on their scores.

\subsection{Single-Agent and Multi-Agent Reasoning}
We employ two representative methods for simulating single-agent and multi-agent reasoning: \textit{Self-Reflection} and \textit{Multi-Agent Debate}:

\paragraph{\textbf{Self-Reflection}}
A single self-reflection agent (\( A_\text{self-reflection} \)) is guided to generate intermediate impact ranking \(\pi\) of CTs and reflection \( R \) alternatively, presenting self-reflection over \( T \) rounds. At each round: 
\[
    (\pi^{t}, R^{t}) = A_\text{self-reflection}(\text{CTs}, \pi^{t-1}).
\]
where \( \pi^t \) and \( R^t \) (\( t \in \{1, 2, \ldots, T\} \)) represent the predicted ranking and reflection on this ranking at round \(t\), respectively. The self-reflection process will stop if there is no further update on the predicted ranking at round \(t\) or if it reaches the pre-defined maximum number of rounds.

\paragraph{\textbf{Multi-Agent Debating}} The multi-agent debating framework includes three main agents: the affirmative debater (\( A_{\text{affirmative}} \)), the negative debater (\( A_{\text{negative}} \)), and the moderator (\( A_{\text{moderator}} \)). The process consists of three phases: Base Solution Generation, Debating, and Judgment.

\textit{Base Solution Generation}: The affirmative debater \( A_{\text{affirmative}} \) is given a list of CTs and generates an initial impact ranking \( \pi_{\text{affirmative}} \):
\[
    \pi_{\text{affirmative}} = A_{\text{affirmative}}(\text{CTs}).
\]

\textit{Debating}: The negative debater \( A_{\text{negative}} \) and the affirmative debater \( A_{\text{affirmative}} \) each receive the list of CTs and the other's proposed solution. They are instructed to disagree and provide their own rankings $\pi_{\text{negative}}$ and $\pi_{\text{affirmative}}$ along with their reasoning $R_{\text{negative}}$ and $R_{\text{affirmative}}$, respectively:

\begin{align}
    (\pi_{\text{negative}}, R_{\text{negative}}) &= A_{\text{negative}}(\text{CTs}, \pi_{\text{affirmative}}), \\
    (\pi_{\text{affirmative}}, R_{\text{affirmative}}) &= A_{\text{affirmative}}(\text{CTs}, \pi_{\text{negative}}).
\end{align}

\textit{Judgment}: The moderator \( A_{\text{moderator}} \) evaluates the arguments and solutions presented by both debaters. They determine if there is a clear preference for one of the solutions. If so, the moderator provides the final impact ranking \( \pi^* \) and summarizes the reasons for supporting either the affirmative or negative side:
\[
    \pi^* = A_{\text{moderator}}(\text{CTs}, \pi_{\text{affirmative}}, \pi_{\text{negative}}, R_{\text{negative}}, R_{\text{affirmative}}).
\]
If no clear preference is established, the debate proceeds to the next round, and the debaters continue to present further arguments.

\section{Experiments}
\subsection{Experimental Setup}
\paragraph{\textbf{Evaluated LLMs.}} We conduct experiments on a set of state-of-the-art proprietary and open-source LLMs to evaluate the efficacy of human-like CT impact assessment using the aforementioned strategies. Specifically, we examine five strong LLMs:
\begin{itemize}
    \item \texttt{GPT-4o-2024-11-20}~\cite{hurst2024gpt}
    \item \texttt{GPT-o1-preview-2024-09-12}~\cite{openai2024o1preview}
    \item \texttt{Llama3.1-70B-Instruct-Turbo}~\cite{dubey2024llama}
    \item \texttt{Qwen2.5-72B-Instruct-Turbo}~\cite{yang2024qwen2}
    \item \texttt{Mixtral-8x22B-Instruct-v0.1}~\cite{mistral2024mixtral}
\end{itemize}

We also evaluate three smaller LLMs::
\begin{itemize}
    \item \texttt{Llama3.1-8b-Instruct-Turbo}~\cite{dubey2024llama}
    \item \texttt{Qwen2.5-7B-Instruct-Turbo}~\cite{yang2024qwen2}
    \item \texttt{Mixtral-7B-Instruct-v0.3}~\cite{mistral2024mixtral}
\end{itemize}

For multi-agent debating, we use two same LLMs as debaters and employ a different LLM as the judge.

\paragraph{\textbf{Evaluation Metrics.}}
We evaluate the performance of different LLMs using the following metrics:
\begin{itemize}
    \item Spearman’s Rank Correlation (\(r_s\))~\cite{spearman1961proof}
    \item Kendall’s Tau (\(\tau\))~\cite{sen1968estimates}
    \item Normalized Discounted Cumulative Gain (\(nDCG\))~\cite{jarvelin2002cumulated}
\end{itemize}

\textit{Spearman’s Rank Correlation} measures the monotonic relationship between predicted and ground truth rankings. Values near 1 or -1 indicate strong positive or negative correlation, respectively. 
The formula is:
\begin{equation}
\label{eq:spearman}
r_s = 1 - \frac{6 \sum d_i^2}{n(n^2 - 1)}
\end{equation}
where \(d_i\) is the difference in rankings for the \(i\)-th item, and \(n\) is the total number of ranked items.

\textit{Kendall’s Tau} evaluates pairwise agreement between predicted and ground truth rankings, with values close to 1 indicating high concordance and -1 indicating strong discordance.
The formula is given by:
\begin{equation}
\label{eq:kendall}
\tau = \frac{(C - D)}{\frac{n(n-1)}{2}}
\end{equation}
where \(C\) is the number of concordant pairs, \(D\) is the number of discordant pairs, and \(n\) is the total number of ranked items.

\textit{Normalized Discounted Cumulative Gain} 
assesses ranking quality by prioritizing higher-ranked items. Scores near 1 indicate strong alignment with the ground truth, with errors at higher positions penalized more heavily.
The formula is:
\begin{equation}
\label{eq:dcg}
DCG = \sum_{i=1}^n \frac{\text{rel}(i)}{\log_2(i+1)}
\end{equation}
where we use \(\text{rel}(i)=1/rank_i\) to assign a relevance score based on the ranking of the \(i\)-th item. Then the nDCG is computed as:

\begin{equation}
\label{eq:ndcg}
nDCG = \frac{DCG}{IDCG}
\end{equation}
where \(IDCG\) is the ideal DCG, calculated from the ground truth ranking. Note that \(nDCG\) penalizes errors more heavily for misrankings at higher positions.

\subsection{Main Results}
\begin{table*}[t]
\centering
\caption{CT impact assessment performance comparison of LLMs across different prompting strategies. Bold values indicate the best results for each metric. All experiments are conducted on the original dataset, and each impact assessment (for every model and prompting strategy) is repeated 3 times, with the average performance reported. *** $p < 0.001$; ** $p < 0.01$; * $p < 0.05$.}
\label{tab:performance-comparison}
\resizebox{\textwidth}{!}{%
\begin{tabular}{|l|ccc|ccc||ccc|ccc||ccc|ccc|}
\hline
{} & \multicolumn{6}{c||}{\textbf{Fast vs. Slow Thinkng}} & \multicolumn{6}{c||}{\textbf{Comparative vs. Scoring Assessment}} & \multicolumn{6}{c|}{\textbf{Single-Agent vs. Multi-Agent Reasoning}} \\ \hline
{} & \multicolumn{3}{c|}{\textbf{Vanilla Ranking}} & \multicolumn{3}{c||}{\textbf{CoT}} & \multicolumn{3}{c|}{\textbf{Scoring}} & \multicolumn{3}{c||}{\textbf{Comparison}} & \multicolumn{3}{c|}{\textbf{Self-Reflection}} & \multicolumn{3}{c|}{\textbf{Debating}} \\ \hline
 & \(r_s\)(\(\uparrow\)) & \(\tau\)(\(\uparrow\)) & \(nDCG\)(\(\uparrow\)) & \(r_s\)(\(\uparrow\)) & \(\tau\)(\(\uparrow\)) & \(nDCG\)(\(\uparrow\)) & \(r_s\)(\(\uparrow\)) & \(\tau\)(\(\uparrow\)) & \(nDCG\)(\(\uparrow\)) & \(r_s\)(\(\uparrow\)) & \(\tau\)(\(\uparrow\)) & \(nDCG\)(\(\uparrow\)) & \(r_s\)(\(\uparrow\)) & \(\tau\)(\(\uparrow\)) & \(nDCG\)(\(\uparrow\)) & \(r_s\)(\(\uparrow\)) & \(\tau\)(\(\uparrow\)) & \(nDCG\)(\(\uparrow\)) \\ \hline
\multicolumn{19}{|c|}{\textbf{Smaller LLMs (<10B parameters)}} \\ \hline
Llama8B & \textbf{-0.02} & \textbf{0.02}* & \textbf{0.67} & 0.02*** & -0.04*** & 0.74 & -0.11 & -0.20* & 0.67 & 0.16 & 0.11 & 0.70 & 0.32 & 0.27* & 0.69 & 0.41** & 0.35** & 0.83 \\ 
Qwen7B & -0.13* & -0.11* & 0.66 & \textbf{0.35} & \textbf{0.32} & \textbf{0.75} & \textbf{0.23} & \textbf{0.19}* & \textbf{0.73} & \textbf{0.51}** & \textbf{0.42}** & 0.73 & \textbf{0.39}* & \textbf{0.33}* & \textbf{0.71} & \textbf{0.62}*** & \textbf{0.44}*** & \textbf{0.85} \\ 
Mistral7B & -0.26 & -0.18 & 0.63 & -0.09 & -0.06 & 0.68 & -0.07 & -0.02 & 0.65 & 0.37* & 0.29* & \textbf{0.79} & -0.51* & -0.45* & 0.59 & 0.52* & 0.39** & 0.84 \\ \hline
\multicolumn{19}{|c|}{\textbf{Larger LLMs (>70B parameters)}} \\ \hline
Llama70B & 0.32 & 0.25* & 0.72 & 0.42 & 0.34* & 0.79 & 0.15 & 0.13 & 0.68 & 0.37 & 0.29* & 0.82 & \textbf{0.60}** & \textbf{0.49}** & \textbf{0.80} & 0.44 & 0.42* & 0.76 \\ 
Qwen72B & 0.39 & 0.29* & 0.78 & 0.49* & 0.40** & 0.79 & 0.35 & 0.28 & 0.74 & 0.59* & 0.50* & 0.79 & 0.57 & 0.42 & 0.75 & \textbf{0.66}** & \textbf{0.52}** & \textbf{0.87} \\ 
Mistral8x22B & -0.53* & -0.42** & 0.58 & 0.44*** & 0.38*** & 0.81 & 0.39 & 0.41 & 0.71 & 0.58** & 0.50** & 0.83 & -0.89*** & -0.72*** & 0.54 & 0.53** & 0.45** & 0.87 \\ 
GPT-4o & 0.42* & 0.35* & 0.83 & 0.52* & 0.44* & 0.82 & 0.40 & 0.35* & 0.78 & 0.55 & 0.46* & 0.81 & -- & -- & -- & -- & -- & -- \\ 
GPT-o1 & \textbf{0.48}* & \textbf{0.40}* & \textbf{0.88} & \textbf{0.60}*** & \textbf{0.50}*** & \textbf{0.86} & \textbf{0.50}** & \textbf{0.44}** & \textbf{0.82} & \textbf{0.62}* & \textbf{0.51}** & \textbf{0.85} & -- & -- & -- & -- & -- & -- \\ \hline
\multicolumn{19}{|c|}{\textbf{Average Performance of LLMs}} \\ \hline
\textbf{Avg. (LLMs<10B)} & -0.14 & -0.09 & 0.65 & 0.09 & 0.07 & 0.72 & 0.02 & -0.01 & 0.68 & 0.35 & 0.27 & 0.74 & 0.07 & 0.05 & 0.66 & 0.52 & 0.39 & \textbf{0.84} \\ 
\textbf{Avg. (LLMs>70B)} & \textbf{0.22} & \textbf{0.17} & \textbf{0.76} & \textbf{0.49} & \textbf{0.42} & \textbf{0.82} & \textbf{0.36} & \textbf{0.30} & \textbf{0.75} & \textbf{0.55} & \textbf{0.46} & \textbf{0.83} & \textbf{0.10} & \textbf{0.19} & \textbf{0.69} & \textbf{0.54} & \textbf{0.46} & \text{0.83} \\ \hline
\end{tabular}%
}
\end{table*}

Table~\ref{tab:performance-comparison} presents the performance of eight evaluated LLMs across three evaluation metrics and five strategies for human-like CT impact assessment. Note that all experimental results in this table come from using the original CT impact assessment dataset (See Appendix for additional experimental results). We separately showcase the performance of smaller and larger LLMs for a better comparison. Among the evaluated models and strategies:

\paragraph{\textbf{Smaller LLMs (<10B parameters).}} 
\textbf{Smaller LLMs generally show poor performance in CT impact assessment}, with most metrics near or below zero for simpler prompting strategies such as Vanilla Ranking and Individual Scoring. A notable exception is the Multi-Agent Debating, where smaller models demonstrate improved performance. For example, \texttt{Qwen7B} stands out as the most consistent among smaller LLMs, achieving strong results in Multi-Agent Debating (\(r_s = 0.62\), \(\tau = 0.44\), \(nDCG = 0.85\)). Moreover, all smaller LLMs benefit from Pairwise Comparison, with significant performance increases.

Another noteworthy observation is the performance of \texttt{Mistral}\\\texttt{7B}, which shows significant improvement in Multi-Agent Debating but decreases with Self-Reflection. One reason could be \texttt{Qwen7B} overly relies on strong LLM to judge their mistakes. It shows very limited self-reflection ability. Furthermore, Multi-Agent Debating emerges as an effective prompting strategy, which significantly improves the performance of all smaller LLMs.

\paragraph{\textbf{Larger LLMs (>70B parameters).}} 
Larger LLMs dominate across all metrics and strategies, significantly outperforming their smaller counterparts. \texttt{GPT-o1} demonstrates exceptional proficiency in CT impact assessment, particularly using CoT (\(r_s = 0.60\), \(\tau = 0.50\), \(nDCG = 0.86\)) and Pairwise Comparison (\(r_s = 0.62\), \(\tau = 0.51\), \(nDCG = 0.85\)). These results also highlight the \texttt{GPT-o1}'s inherent ability to handle complex tasks that require step-by-step analyses and nuanced reasoning. Other larger LLMs, such as \texttt{Qwen72B} and \texttt{Mistral8x22B}, also perform strongly in Pairwise Comparison, with \texttt{Mistral8x22B} achieving the second best \(nDCG = 0.83\). However, there are slight variations among the larger LLMs. For example, \texttt{Llama70B} shows relatively lower performance in Vanilla Ranking and Individual Scoring. One anomaly case is \texttt{Mistral8x22B}, which has (\(r_s = -0.53\) and -0.89 using Vanilla Ranking and Self-Reflection, respectively. 

On average, \textbf{larger LLMs achieve significantly better results across all metrics and prompting strategies}, especially with CoT. The average performance of larger LLMs (\(Avg.\; r_s = 0.49\), \(Avg.\; \tau = 0.42\), \(Avg.\; nDCG = 0.82\)) far exceeds that of smaller LLMs (\(Avg.\; r_s = 0.09\), \(Avg.\; \tau = 0.07\), \(Avg.\; nDCG = 0.72\)). Furthermore, advanced strategies like Self-Reflection and Debating, allow these models to capitalize on their extensive knowledge bases for CT impact assessment.

\subsection{Impact of Different Strategies}
As the results demonstrated in Table~\ref{tab:performance-comparison}, different strategies significantly affect the efficacy of LLMs in CT impact assessment. By simulating various human-like reasoning processes (fast and slow thinking), assessment paradigms (comparison and scoring), and interactive behaviors (single-agent reasoning and multi-agent debating), we systematically evaluate how different strategies affect CT impact assessment.

\paragraph{\textbf{Fast and Slow Thinking.}}  
The comparison between Vanilla Ranking and CoT reveals the impact of simulating fast versus slow thinking. For smaller LLMs, both Vanilla Ranking and CoT perform poorly with average \(r_s\) and \(\tau\) close to zero, which indicates a limited capacity for instinctive CT impact assessment. However, \texttt{Qwen7B} benefit from slow thinking a lot with \(r_s\) and \(\tau\) increase from negative to 0.35 and 0.32, respectively. On the other hand, larger LLMs exhibit good performance with Vanilla Ranking and CoT, suggesting that these models inherently integrate sufficient reasoning ability even without explicit guidance. In some cases, Vanilla Ranking even outperforms CoT in certain metrics, as observed with \texttt{GPT-4o} and \texttt{GPT-o1}. These findings suggest that \textbf{slow thinking mode improves LLM performance in CT impact assessment, particularly for smaller LLMs}. However, for larger LLMs, the thinking mode appears to be less important, as they consistently achieve good performance regardless of whether fast or slow thinking is employed.

\begin{figure*}[!h]
    \centering
    \begin{subfigure}{\columnwidth}
        \includegraphics[width=1\columnwidth]{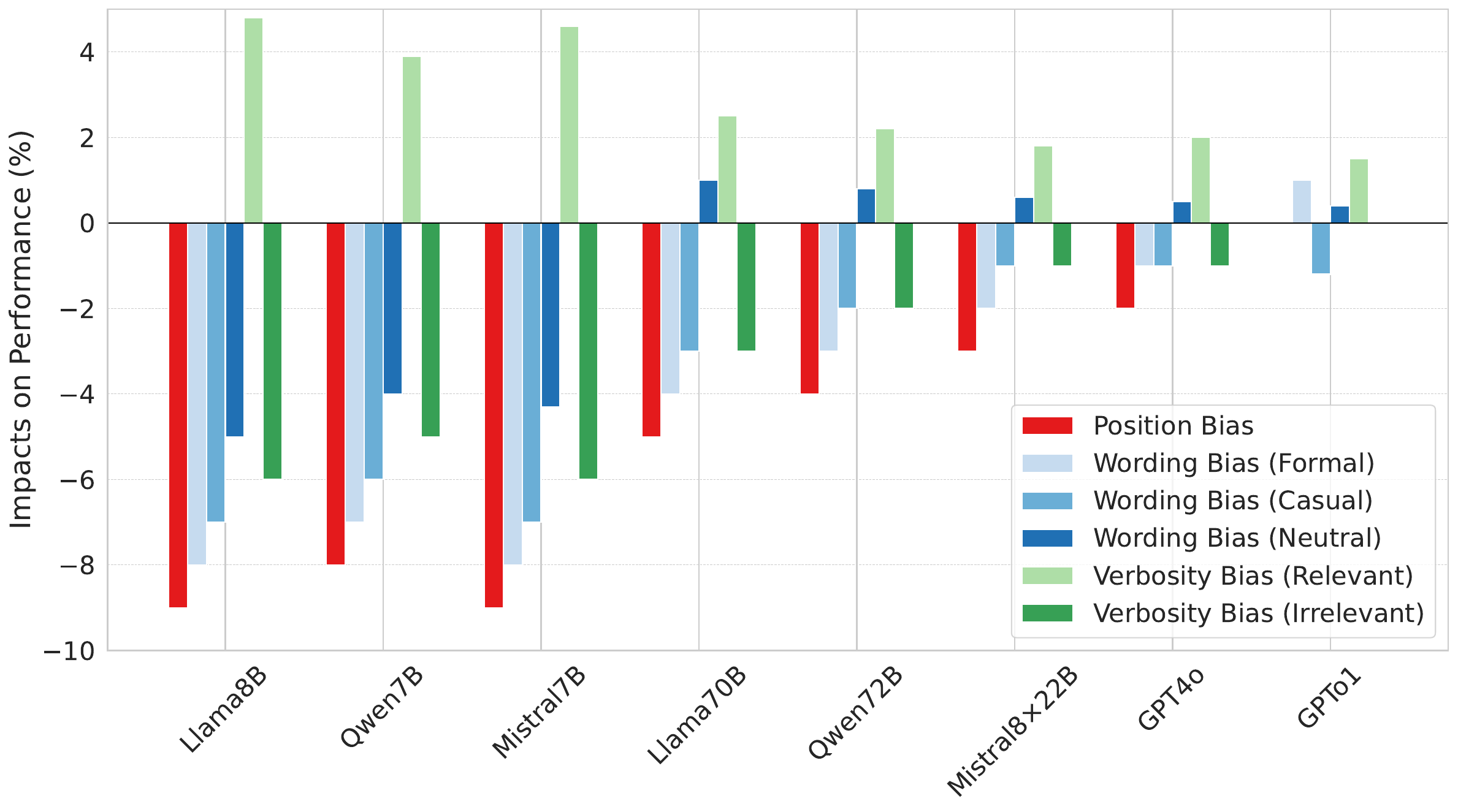}
        \caption{Vanilla Ranking.}
        \label{fig:Vanilla}
    \end{subfigure}
    \begin{subfigure}{\columnwidth}
        \includegraphics[width=1\columnwidth]{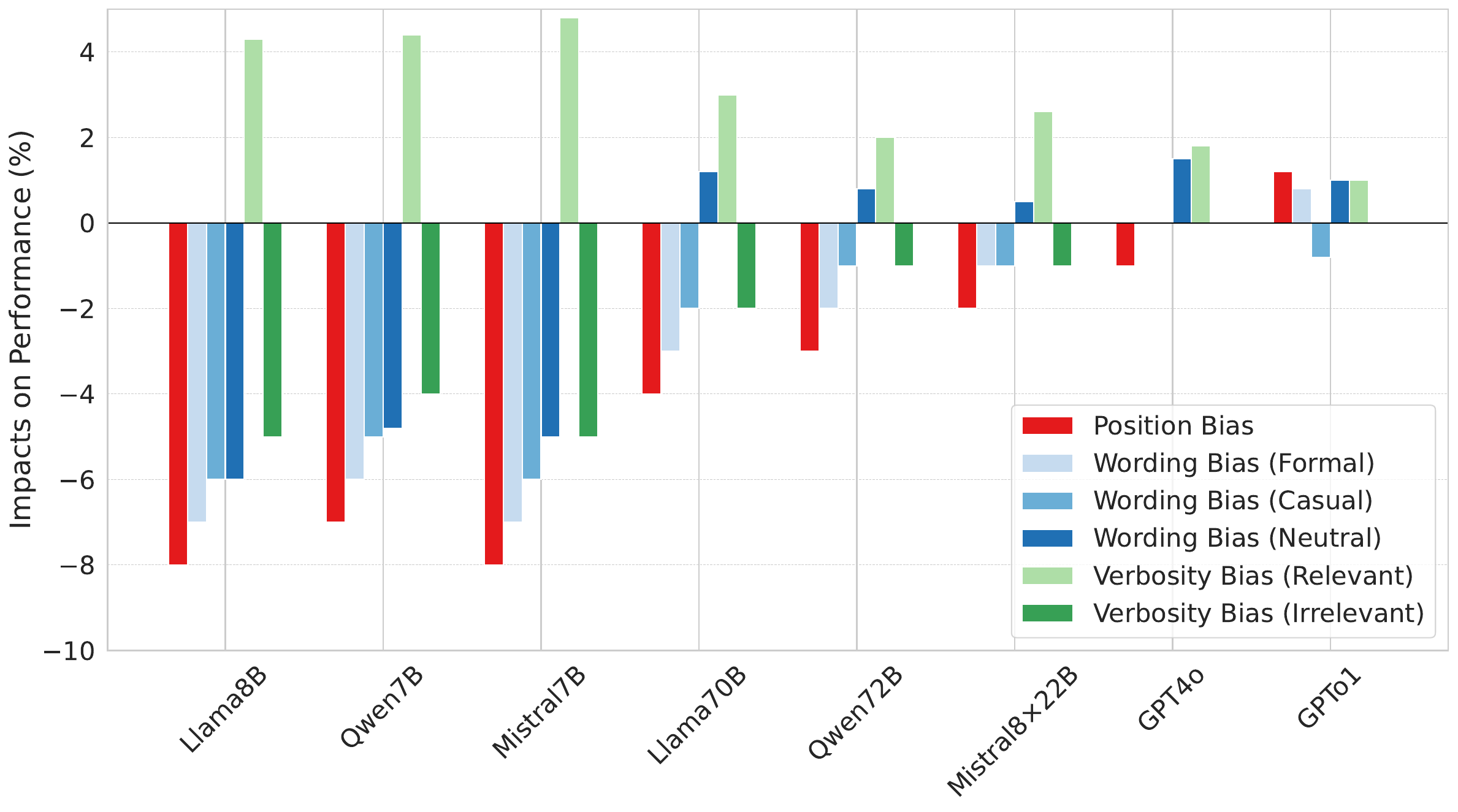}
        \caption{Individual Scoring.}
        \label{fig:Scoring}
    \end{subfigure}
    \begin{subfigure}{\columnwidth}
        \includegraphics[width=1\columnwidth]{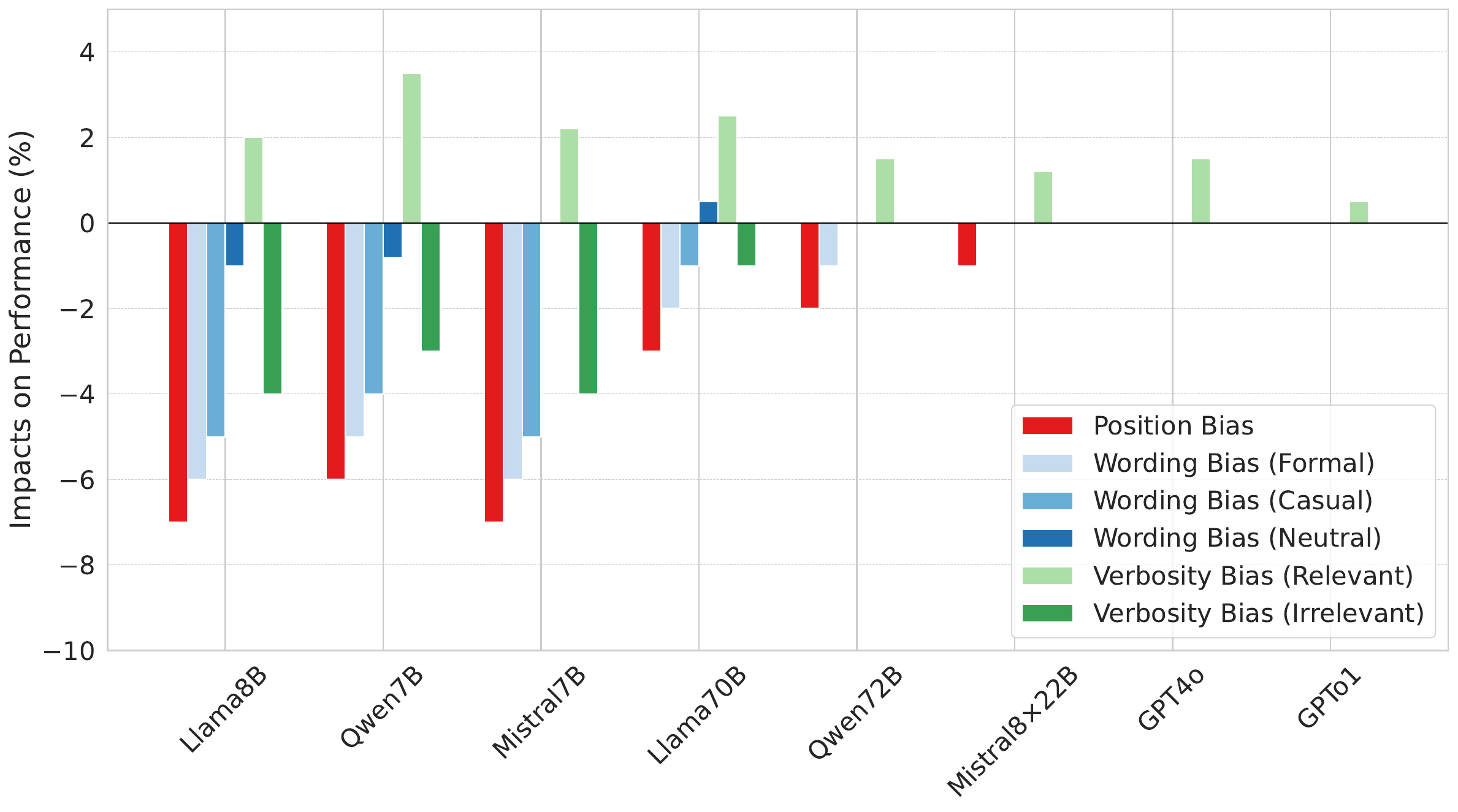}
        \caption{Pairwise Comparison.}
        \label{fig:Comparison}
    \end{subfigure}
    \begin{subfigure}{\columnwidth}
        \includegraphics[width=1\columnwidth]{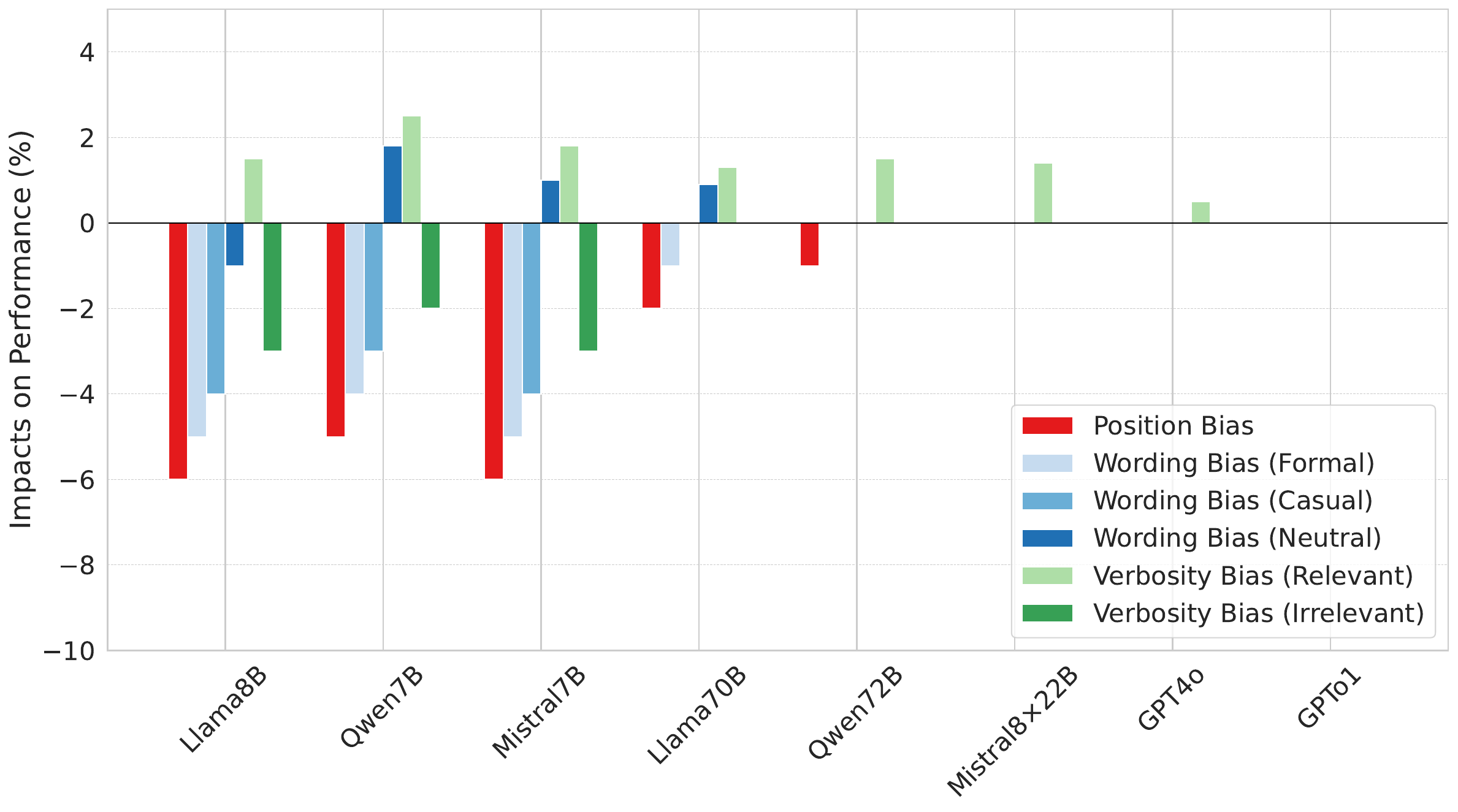}
        \caption{Chain-of-Thought.}
        \label{fig:CoT}
    \end{subfigure}
    \caption{Impacts ($\tau$) of prompting biases on LLM performance across different evaluation strategies: (a) Vanilla Ranking, (b) Individual Scoring, (c) Pairwise Comparison, and (d) Chain-of-Thought. The bar plots show the relative performance changes introduced by position bias, wording bias (formal, casual, neutral), and verbosity bias (relevant and irrelevant). Results of Multi-Agent Debating are not presented here as they show negligible or no performance changes across all bias categories.}

    \label{fig:biases}
\end{figure*}

\paragraph{\textbf{Comparative and Scoring Assessment.}}  
The choice between Pairwise Comparison and Individual Scoring paradigms can significantly affect the CT impact assessment performance. We observe that \textbf{Pairwise Comparison consistently outperforms Individual Scoring across all LLMs}, which suggests that LLMs are more adept at step-by-step comparison to produce accurate CT impact rankings. Smaller LLMs, such as \texttt{Qwen7B}, achieve \(r_s = 0.51\), \(\tau=0.42\) and \(nDCG = 0.73\) under Pairwise Comparison, outperforming their scores in Individual Scoring (\(r_s = 0.23\), \(\tau=0.19\), and \(nDCG = 0.73\)). Larger LLMs also benefit from Pairwise Comparison, with \texttt{GPT-o1} achieving the best \(r_s = 0.62\) and \(\tau=0.51\) across all LLMs.

Individual Scoring requires LLMs to assign an impact score for each CT, which may potentially be more prone to calibration errors and inconsistencies. CT impact assessments from smaller LLMs demonstrate poor correlations with human annotations in Individual Scoring (\(Avg.\; r_s = 0.02\) and \(Avg.\; \tau = -0.01\)). However, while larger LLMs maintain reasonable performance, we speculate this is due to their larger training corpora. Compared to larger LLM's performance using Vanilla Ranking, using Individual Scoring gets similar or even worse performance (see \texttt{Llama70B}).

\paragraph{\textbf{Single-Agent Reasoning and Multi-Agent Debating.}}  
The comparison between Single-Agent Reasoning and Multi-Agent Debating highlights the importance of interactive dynamics in CT impact assessment. Self-Reflection leverages a single LLM agent to repeatedly correct and improve the CT impact assessment. LLMs utilizing this prompting strategy show moderate performance improvements. All LLMs, except the \texttt{Mistral} family, achieve better performance than simpler strategies such as Vanilla Ranking and Individual Scoring. \textbf{We speculate that instead of ``self-reflection'', \texttt{Mistral} experienced an ``error-reinforcement''}, which went further toward the opposite direction through the iterative process. 

On the other hand, Multi-Agent Debating introduces collaborative and adversarial interactions among LLMs, significantly enhancing performance for all smaller LLMs. By combining diverse perspectives and iterative analyses, \textbf{Multi-Agent Debating enables smaller LLMs to provide accurate CT impact assessment, matching or exceeding the performance of larger LLMs}. Note that in Multi-Agent Debating, we use the smaller LLMs as debaters (e.g., \texttt{Llama8B}) and their larger counterparts as a judge (e.g., Llama70B). We only use larger LLMs to select a ranking generated by smaller LLMs. In other words, larger LLMs are not allowed to provide new CT impact rankings by themselves. We didn't apply Self-Reflection and Multi-Agent Debating to \texttt{GPT-4o} and \texttt{GPT-o1} because of the expensive costs. We conclude that \textbf{Multi-Agent Debating is the most effective strategy for CT impact assessment}, which enables LLMs to think critically and systematically analyze more CT-related information.

\subsection{Impact of Prompting Biases}
Analyzing prompting bias impacts on CT impact assessment reveals vulnerabilities across different strategies. Figure~\ref{fig:biases} illustrates the comparative performance variations resulting from position, wording, and verbosity biases under various prompting strategies. Robustness testing results of Self-Reflection and Multi-Agent Debating are omitted due to negligible performance (close to zero) changes across all prompting bias categories.

\subsubsection{\textbf{Position Bias}}
Position bias consistently causes performance degradation across all strategies, particularly in Vanilla Ranking (Figure~\ref{fig:Vanilla}) and Individual Scoring (Figure~\ref{fig:Scoring}). Smaller LLMs are more affected, showing significant performance drops, while larger LLMs demonstrate greater robustness. Pairwise Comparison (Figure~\ref{fig:Comparison}) and CoT (Figure~\ref{fig:CoT}) demonstrate improved robustness, especially for larger LLMs. We observe that the primary reason for this degradation is that LLMs tend to disproportionately assign higher rankings to CTs listed earlier in the input sequence. In the original dataset, CTs are ordered by ground-truth impact levels. Therefore, the positional dependency can coincidentally make LLMs’ CT impact assessment ``look better'' under a specific order. 

\paragraph{\textbf{Potential Solutions.}} To mitigate position bias, one solution is to use \textbf{randomized position prompts} during inference to break the correlation between input order and LLM predictions. Moreover, \textbf{shuffling the dataset} beforehand can help reduce the model's dependency on positional cues. However, they may not fundamentally address the issue. It would be better to consider \textbf{introducing positional randomness during fine-tuning} to encourage the model to rely less on order and more on semantic content.

\subsubsection{\textbf{Wording Bias}}
Emotionally charged wording (e.g., formal and casual) tends to impact performance more significantly than neutral wording across prompting strategies. Intuitively, rephrasing CTs in different tones, especially a neutral tone, should not influence LLMs' CT impact assessments. We speculate that LLMs leverage implicit linguistic cues for assessing the impact, particularly for smaller LLMs with limited contextual adaptability. Casual phrasing generally leads to the largest performance deviations in smaller models, as shown in Figures~\ref{fig:Vanilla} and \ref{fig:Scoring}. On the other hand, larger LLMs exhibit greater consistency. For example, Pairwise Comparison (Figure~\ref{fig:Comparison}) and CoT (Figure~\ref{fig:CoT}) significantly mitigate this bias by implementing refined analyses rather than token-level inferences. CoT demonstrates the greatest robustness to wording bias, which uses intermediate reasoning steps to neutralize the effects of stylistic differences.

\paragraph{\textbf{Potential Solutions.}} incorporating various \textbf{rephrasing during the prompting stage} is promising. However, while rephrasing can introduce variation and help counteract specific biases, it is inherently reactive. It does not prevent the model from being affected by wording bias in unseen prompts. Moreover, rephrasing may add complexity to the inference stage and unintentionally introduce new biases or inconsistencies.

\subsubsection{\textbf{Verbosity Bias}}
Verbosity bias significantly affects performance when irrelevant content is added to the original CT descriptions. This degradation is most affected in Vanilla Ranking (Figure~\ref{fig:Vanilla}) and Individual Scoring (Figure~\ref{fig:Scoring}), where both smaller and larger LLMs struggle to filter out extraneous information. Smaller LLMs are particularly vulnerable, as they may lack the ability to discern between relevant and irrelevant verbosity. In contrast, larger LLMs seem more robust but still show some performance drops. Pairwise Comparison (Figure~\ref{fig:Comparison}) and CoT (Figure~\ref{fig:CoT}) are notably robust to irrelevant verbosity. In particular, CoT enables intermediate reasoning steps for LLMs to focus on the important aspects of CTS. Interestingly, relevant verbosity has positive effects on performance. One explanation could be it provides additional meaningful context that LLMs can integrate into their analyses. For example, including event timing and target population in prompts provides LLMs with useful contextual information, facilitating thorough assessments. 

\paragraph{\textbf{Potential Solutions.}} A promising approach is to apply preprocessing techniques that \textbf{condense verbose inputs into concise and relevant summaries}. However, there is a risk of inadvertently manipulating CT-related information, causing changes in the perceived impact. Besides, prompting LLMs for \textbf{multi-step reasoning and analysis} can enhance their ability to disregard irrelevant verbosity and prioritize useful information. However, they may not be feasible for large-scale applications.


\section{Limitations and Future Work}
This study has several limitations that future work should consider. First, the datasets used in this work focus on CTs with significant public visibility and media exposure in the US. This could bring cultural and geographical biases, limiting the generalizability of the findings to other regions or less prominent CTs. Moreover, we did not consider newly emerging CTs due to the lack of human annotations, which are critical in real-world applications. Second, the performance of smaller LLMs highlights their limited capability for effectively assessing CT impacts. Larger LLMs demonstrate superior performance but may have challenges in scalability due to high computational costs. Although Multi-Agent Debating seems promising, future work should explore more advanced prompting strategies to balance performance with resource constraints. Last but not least, this study focuses on prompting biases like position, wording, and verbosity. However, interactions between these biases (e.g., how verbosity bias might amplify wording bias) remain unexplored. Other biases, such as political, demographic, and recency, should also be considered.

\section{Conclusions and Discussion}
This study investigates the feasibility of using LLMs for human-like CT impact assessment. By evaluating eight LLMs across five prompting strategies, we conduct extensive experiments to showcase LLMs' general CT impact assessment capability and robustness to different biases. The findings show that larger LLMs outperform smaller models across all evaluation metrics, with Multi-Agent Debating significantly enhancing robustness and accuracy. However, position, wording, and verbosity biases present distinct challenges. To address these challenges, we provide practical mitigation solutions. Last but not least, we discuss our limitations and directions for further research. We suggest designing advanced impact assessment frameworks and ensemble methods that leverage the strengths of vast information extraction and multi-step analysis. Expanding our datasets to include diverse CTs is also critical. 

In conclusion, LLMs guided with tailored prompting strategies appear to be an effective tool for CT impact assessment. By carefully addressing their limitations and biases, these models can be useful in facilitating human-like CT impact assessment.


\bibliographystyle{ACM-Reference-Format}
\bibliography{sample-base}


\begin{thebibliography}{70}


\ifx \showCODEN    \undefined \def \showCODEN     #1{\unskip}     \fi
\ifx \showDOI      \undefined \def \showDOI       #1{#1}\fi
\ifx \showISBNx    \undefined \def \showISBNx     #1{\unskip}     \fi
\ifx \showISBNxiii \undefined \def \showISBNxiii  #1{\unskip}     \fi
\ifx \showISSN     \undefined \def \showISSN      #1{\unskip}     \fi
\ifx \showLCCN     \undefined \def \showLCCN      #1{\unskip}     \fi
\ifx \shownote     \undefined \def \shownote      #1{#1}          \fi
\ifx \showarticletitle \undefined \def \showarticletitle #1{#1}   \fi
\ifx \showURL      \undefined \def \showURL       {\relax}        \fi
\providecommand\bibfield[2]{#2}
\providecommand\bibinfo[2]{#2}
\providecommand\natexlab[1]{#1}
\providecommand\showeprint[2][]{arXiv:#2}

\bibitem[Achiam et~al\mbox{.}(2023)]%
        {achiam2023gpt}
\bibfield{author}{\bibinfo{person}{Josh Achiam}, \bibinfo{person}{Steven Adler}, \bibinfo{person}{Sandhini Agarwal}, \bibinfo{person}{Lama Ahmad}, \bibinfo{person}{Ilge Akkaya}, \bibinfo{person}{Florencia~Leoni Aleman}, \bibinfo{person}{Diogo Almeida}, \bibinfo{person}{Janko Altenschmidt}, \bibinfo{person}{Sam Altman}, \bibinfo{person}{Shyamal Anadkat}, {et~al\mbox{.}}} \bibinfo{year}{2023}\natexlab{}.
\newblock \showarticletitle{Gpt-4 technical report}.
\newblock \bibinfo{journal}{\emph{arXiv preprint arXiv:2303.08774}} (\bibinfo{year}{2023}).
\newblock


\bibitem[AI(2024)]%
        {mistral2024mixtral}
\bibfield{author}{\bibinfo{person}{Mistral AI}.} \bibinfo{year}{2024}\natexlab{}.
\newblock \bibinfo{title}{Mixtral: 8×22B Mixture of Experts}.
\newblock \bibinfo{howpublished}{\url{https://mistral.ai/news/mixtral-8x22b/}}.
\newblock
\newblock
\shownote{Accessed: 2024-11-29}.


\bibitem[Assessment(1995)]%
        {assessment1995guidelines}
\bibfield{author}{\bibinfo{person}{Social~Impact Assessment}.} \bibinfo{year}{1995}\natexlab{}.
\newblock \showarticletitle{Guidelines and principles for social impact assessment}.
\newblock \bibinfo{journal}{\emph{Environmental Impact Assessment Review}} \bibinfo{volume}{15}, \bibinfo{number}{1} (\bibinfo{year}{1995}), \bibinfo{pages}{11--43}.
\newblock


\bibitem[Bai et~al\mbox{.}(2024)]%
        {bai2024benchmarking}
\bibfield{author}{\bibinfo{person}{Yushi Bai}, \bibinfo{person}{Jiahao Ying}, \bibinfo{person}{Yixin Cao}, \bibinfo{person}{Xin Lv}, \bibinfo{person}{Yuze He}, \bibinfo{person}{Xiaozhi Wang}, \bibinfo{person}{Jifan Yu}, \bibinfo{person}{Kaisheng Zeng}, \bibinfo{person}{Yijia Xiao}, \bibinfo{person}{Haozhe Lyu}, {et~al\mbox{.}}} \bibinfo{year}{2024}\natexlab{}.
\newblock \showarticletitle{Benchmarking foundation models with language-model-as-an-examiner}.
\newblock \bibinfo{journal}{\emph{Advances in Neural Information Processing Systems}}  \bibinfo{volume}{36} (\bibinfo{year}{2024}).
\newblock


\bibitem[Bavel et~al\mbox{.}(2020)]%
        {bavel2020using}
\bibfield{author}{\bibinfo{person}{Jay J~Van Bavel}, \bibinfo{person}{Katherine Baicker}, \bibinfo{person}{Paulo~S Boggio}, \bibinfo{person}{Valerio Capraro}, \bibinfo{person}{Aleksandra Cichocka}, \bibinfo{person}{Mina Cikara}, \bibinfo{person}{Molly~J Crockett}, \bibinfo{person}{Alia~J Crum}, \bibinfo{person}{Karen~M Douglas}, \bibinfo{person}{James~N Druckman}, {et~al\mbox{.}}} \bibinfo{year}{2020}\natexlab{}.
\newblock \showarticletitle{Using social and behavioural science to support COVID-19 pandemic response}.
\newblock \bibinfo{journal}{\emph{Nature human behaviour}} \bibinfo{volume}{4}, \bibinfo{number}{5} (\bibinfo{year}{2020}), \bibinfo{pages}{460--471}.
\newblock


\bibitem[Beigi et~al\mbox{.}(2024)]%
        {beigi2024lrq}
\bibfield{author}{\bibinfo{person}{Alimohammad Beigi}, \bibinfo{person}{Bohan Jiang}, \bibinfo{person}{Dawei Li}, \bibinfo{person}{Tharindu Kumarage}, \bibinfo{person}{Zhen Tan}, \bibinfo{person}{Pouya Shaeri}, {and} \bibinfo{person}{Huan Liu}.} \bibinfo{year}{2024}\natexlab{}.
\newblock \showarticletitle{Lrq-fact: Llm-generated relevant questions for multimodal fact-checking}.
\newblock \bibinfo{journal}{\emph{arXiv preprint arXiv:2410.04616}} (\bibinfo{year}{2024}).
\newblock


\bibitem[Brehm(1966)]%
        {brehm1966theory}
\bibfield{author}{\bibinfo{person}{Jack~W Brehm}.} \bibinfo{year}{1966}\natexlab{}.
\newblock \showarticletitle{A theory of psychological reactance.}
\newblock  (\bibinfo{year}{1966}).
\newblock


\bibitem[Cassam(2019)]%
        {cassam2019conspiracy}
\bibfield{author}{\bibinfo{person}{Quassim Cassam}.} \bibinfo{year}{2019}\natexlab{}.
\newblock \bibinfo{booktitle}{\emph{Conspiracy theories}}.
\newblock \bibinfo{publisher}{John Wiley \& Sons}.
\newblock


\bibitem[Chen and Shu(2023)]%
        {chen2023can}
\bibfield{author}{\bibinfo{person}{Canyu Chen} {and} \bibinfo{person}{Kai Shu}.} \bibinfo{year}{2023}\natexlab{}.
\newblock \showarticletitle{Can llm-generated misinformation be detected?}
\newblock \bibinfo{journal}{\emph{arXiv preprint arXiv:2309.13788}} (\bibinfo{year}{2023}).
\newblock


\bibitem[Chen et~al\mbox{.}(2020)]%
        {chen2020knowledge}
\bibfield{author}{\bibinfo{person}{Ying Chen}, \bibinfo{person}{Rui Zhou}, \bibinfo{person}{Boyan Chen}, \bibinfo{person}{Hao Chen}, \bibinfo{person}{Ying Li}, \bibinfo{person}{Zhi Chen}, \bibinfo{person}{Haihong Zhu}, {and} \bibinfo{person}{Hongmei Wang}.} \bibinfo{year}{2020}\natexlab{}.
\newblock \showarticletitle{Knowledge, perceived beliefs, and preventive behaviors related to COVID-19 among Chinese older adults: cross-sectional web-based survey}.
\newblock \bibinfo{journal}{\emph{Journal of Medical Internet Research}} \bibinfo{volume}{22}, \bibinfo{number}{12} (\bibinfo{year}{2020}), \bibinfo{pages}{e23729}.
\newblock


\bibitem[Daniel(2017)]%
        {daniel2017thinking}
\bibfield{author}{\bibinfo{person}{Kahneman Daniel}.} \bibinfo{year}{2017}\natexlab{}.
\newblock \bibinfo{booktitle}{\emph{Thinking, fast and slow}}.
\newblock


\bibitem[Diab et~al\mbox{.}(2024)]%
        {diab2024classifying}
\bibfield{author}{\bibinfo{person}{Ahmad Diab}, \bibinfo{person}{Rr Nefriana}, {and} \bibinfo{person}{Yu-Ru Lin}.} \bibinfo{year}{2024}\natexlab{}.
\newblock \showarticletitle{Classifying Conspiratorial Narratives at Scale: False Alarms and Erroneous Connections}. In \bibinfo{booktitle}{\emph{Proceedings of the International AAAI Conference on Web and Social Media}}, Vol.~\bibinfo{volume}{18}. \bibinfo{pages}{340--353}.
\newblock


\bibitem[Douglas et~al\mbox{.}(2017)]%
        {douglas2017psychology}
\bibfield{author}{\bibinfo{person}{Karen~M Douglas}, \bibinfo{person}{Robbie~M Sutton}, {and} \bibinfo{person}{Aleksandra Cichocka}.} \bibinfo{year}{2017}\natexlab{}.
\newblock \showarticletitle{The psychology of conspiracy theories}.
\newblock \bibinfo{journal}{\emph{Current directions in psychological science}} \bibinfo{volume}{26}, \bibinfo{number}{6} (\bibinfo{year}{2017}), \bibinfo{pages}{538--542}.
\newblock


\bibitem[Douglas et~al\mbox{.}(2019)]%
        {douglas2019understanding}
\bibfield{author}{\bibinfo{person}{Karen~M Douglas}, \bibinfo{person}{Joseph~E Uscinski}, \bibinfo{person}{Robbie~M Sutton}, \bibinfo{person}{Aleksandra Cichocka}, \bibinfo{person}{Turkay Nefes}, \bibinfo{person}{Chee~Siang Ang}, {and} \bibinfo{person}{Farzin Deravi}.} \bibinfo{year}{2019}\natexlab{}.
\newblock \showarticletitle{Understanding conspiracy theories}.
\newblock \bibinfo{journal}{\emph{Political psychology}}  \bibinfo{volume}{40} (\bibinfo{year}{2019}), \bibinfo{pages}{3--35}.
\newblock


\bibitem[Dubey et~al\mbox{.}(2024)]%
        {dubey2024llama}
\bibfield{author}{\bibinfo{person}{Abhimanyu Dubey}, \bibinfo{person}{Abhinav Jauhri}, \bibinfo{person}{Abhinav Pandey}, \bibinfo{person}{Abhishek Kadian}, \bibinfo{person}{Ahmad Al-Dahle}, \bibinfo{person}{Aiesha Letman}, \bibinfo{person}{Akhil Mathur}, \bibinfo{person}{Alan Schelten}, \bibinfo{person}{Amy Yang}, \bibinfo{person}{Angela Fan}, {et~al\mbox{.}}} \bibinfo{year}{2024}\natexlab{}.
\newblock \showarticletitle{The llama 3 herd of models}.
\newblock \bibinfo{journal}{\emph{arXiv preprint arXiv:2407.21783}} (\bibinfo{year}{2024}).
\newblock


\bibitem[Enders et~al\mbox{.}(2022)]%
        {enders2022relationship}
\bibfield{author}{\bibinfo{person}{Adam~M Enders}, \bibinfo{person}{Joseph Uscinski}, \bibinfo{person}{Casey Klofstad}, {and} \bibinfo{person}{Justin Stoler}.} \bibinfo{year}{2022}\natexlab{}.
\newblock \showarticletitle{On the relationship between conspiracy theory beliefs, misinformation, and vaccine hesitancy}.
\newblock \bibinfo{journal}{\emph{Plos one}} \bibinfo{volume}{17}, \bibinfo{number}{10} (\bibinfo{year}{2022}), \bibinfo{pages}{e0276082}.
\newblock


\bibitem[Esteves et~al\mbox{.}(2012)]%
        {esteves2012social}
\bibfield{author}{\bibinfo{person}{Ana~Maria Esteves}, \bibinfo{person}{Daniel Franks}, {and} \bibinfo{person}{Frank Vanclay}.} \bibinfo{year}{2012}\natexlab{}.
\newblock \showarticletitle{Social impact assessment: the state of the art}.
\newblock \bibinfo{journal}{\emph{Impact assessment and project appraisal}} \bibinfo{volume}{30}, \bibinfo{number}{1} (\bibinfo{year}{2012}), \bibinfo{pages}{34--42}.
\newblock


\bibitem[Freeman et~al\mbox{.}(2022)]%
        {freeman2022coronavirus}
\bibfield{author}{\bibinfo{person}{Daniel Freeman}, \bibinfo{person}{Felicity Waite}, \bibinfo{person}{Laina Rosebrock}, \bibinfo{person}{Ariane Petit}, \bibinfo{person}{Chiara Causier}, \bibinfo{person}{Anna East}, \bibinfo{person}{Lucy Jenner}, \bibinfo{person}{Ashley-Louise Teale}, \bibinfo{person}{Lydia Carr}, \bibinfo{person}{Sophie Mulhall}, {et~al\mbox{.}}} \bibinfo{year}{2022}\natexlab{}.
\newblock \showarticletitle{Coronavirus conspiracy beliefs, mistrust, and compliance with government guidelines in England}.
\newblock \bibinfo{journal}{\emph{Psychological medicine}} \bibinfo{volume}{52}, \bibinfo{number}{2} (\bibinfo{year}{2022}), \bibinfo{pages}{251--263}.
\newblock


\bibitem[Gallacher et~al\mbox{.}(2021)]%
        {gallacher2021online}
\bibfield{author}{\bibinfo{person}{John~D Gallacher}, \bibinfo{person}{Marc~W Heerdink}, {and} \bibinfo{person}{Miles Hewstone}.} \bibinfo{year}{2021}\natexlab{}.
\newblock \showarticletitle{Online engagement between opposing political protest groups via social media is linked to physical violence of offline encounters}.
\newblock \bibinfo{journal}{\emph{Social Media+ Society}} \bibinfo{volume}{7}, \bibinfo{number}{1} (\bibinfo{year}{2021}), \bibinfo{pages}{2056305120984445}.
\newblock


\bibitem[Gao et~al\mbox{.}(2023a)]%
        {gao2023s}
\bibfield{author}{\bibinfo{person}{Chen Gao}, \bibinfo{person}{Xiaochong Lan}, \bibinfo{person}{Zhihong Lu}, \bibinfo{person}{Jinzhu Mao}, \bibinfo{person}{Jinghua Piao}, \bibinfo{person}{Huandong Wang}, \bibinfo{person}{Depeng Jin}, {and} \bibinfo{person}{Yong Li}.} \bibinfo{year}{2023}\natexlab{a}.
\newblock \showarticletitle{S3: Social-network Simulation System with Large Language Model-Empowered Agents}.
\newblock \bibinfo{journal}{\emph{arXiv preprint arXiv:2307.14984}} (\bibinfo{year}{2023}).
\newblock


\bibitem[Gao et~al\mbox{.}(2024)]%
        {gao2024taxonomy}
\bibfield{author}{\bibinfo{person}{Jie Gao}, \bibinfo{person}{Simret~Araya Gebreegziabher}, \bibinfo{person}{Kenny Tsu~Wei Choo}, \bibinfo{person}{Toby Jia-Jun Li}, \bibinfo{person}{Simon~Tangi Perrault}, {and} \bibinfo{person}{Thomas~W Malone}.} \bibinfo{year}{2024}\natexlab{}.
\newblock \showarticletitle{A Taxonomy for Human-LLM Interaction Modes: An Initial Exploration}. In \bibinfo{booktitle}{\emph{Extended Abstracts of the CHI Conference on Human Factors in Computing Systems}}. \bibinfo{pages}{1--11}.
\newblock


\bibitem[Gao et~al\mbox{.}(2023b)]%
        {gao2023human}
\bibfield{author}{\bibinfo{person}{Mingqi Gao}, \bibinfo{person}{Jie Ruan}, \bibinfo{person}{Renliang Sun}, \bibinfo{person}{Xunjian Yin}, \bibinfo{person}{Shiping Yang}, {and} \bibinfo{person}{Xiaojun Wan}.} \bibinfo{year}{2023}\natexlab{b}.
\newblock \showarticletitle{Human-like summarization evaluation with chatgpt}.
\newblock \bibinfo{journal}{\emph{arXiv preprint arXiv:2304.02554}} (\bibinfo{year}{2023}).
\newblock


\bibitem[Goertzel(1994)]%
        {goertzel1994belief}
\bibfield{author}{\bibinfo{person}{Ted Goertzel}.} \bibinfo{year}{1994}\natexlab{}.
\newblock \showarticletitle{Belief in conspiracy theories}.
\newblock \bibinfo{journal}{\emph{Political psychology}} (\bibinfo{year}{1994}), \bibinfo{pages}{731--742}.
\newblock


\bibitem[Goodman and Carmichael(2020)]%
        {goodman2020coronavirus}
\bibfield{author}{\bibinfo{person}{Jack Goodman} {and} \bibinfo{person}{Flora Carmichael}.} \bibinfo{year}{2020}\natexlab{}.
\newblock \showarticletitle{Coronavirus: Bill Gates ‘microchip’conspiracy theory and other vaccine claims fact-checked}.
\newblock \bibinfo{journal}{\emph{BBC News}}  \bibinfo{volume}{30} (\bibinfo{year}{2020}).
\newblock


\bibitem[Huang et~al\mbox{.}(2024)]%
        {huang2024social}
\bibfield{author}{\bibinfo{person}{Yue Huang}, \bibinfo{person}{Zhengqing Yuan}, \bibinfo{person}{Yujun Zhou}, \bibinfo{person}{Kehan Guo}, \bibinfo{person}{Xiangqi Wang}, \bibinfo{person}{Haomin Zhuang}, \bibinfo{person}{Weixiang Sun}, \bibinfo{person}{Lichao Sun}, \bibinfo{person}{Jindong Wang}, \bibinfo{person}{Yanfang Ye}, {et~al\mbox{.}}} \bibinfo{year}{2024}\natexlab{}.
\newblock \showarticletitle{Social Science Meets LLMs: How Reliable Are Large Language Models in Social Simulations?}
\newblock \bibinfo{journal}{\emph{arXiv preprint arXiv:2410.23426}} (\bibinfo{year}{2024}).
\newblock


\bibitem[Hurst et~al\mbox{.}(2024)]%
        {hurst2024gpt}
\bibfield{author}{\bibinfo{person}{Aaron Hurst}, \bibinfo{person}{Adam Lerer}, \bibinfo{person}{Adam~P Goucher}, \bibinfo{person}{Adam Perelman}, \bibinfo{person}{Aditya Ramesh}, \bibinfo{person}{Aidan Clark}, \bibinfo{person}{AJ Ostrow}, \bibinfo{person}{Akila Welihinda}, \bibinfo{person}{Alan Hayes}, \bibinfo{person}{Alec Radford}, {et~al\mbox{.}}} \bibinfo{year}{2024}\natexlab{}.
\newblock \showarticletitle{Gpt-4o system card}.
\newblock \bibinfo{journal}{\emph{arXiv preprint arXiv:2410.21276}} (\bibinfo{year}{2024}).
\newblock


\bibitem[Innes and Innes(2023)]%
        {innes2023platforming}
\bibfield{author}{\bibinfo{person}{Helen Innes} {and} \bibinfo{person}{Martin Innes}.} \bibinfo{year}{2023}\natexlab{}.
\newblock \showarticletitle{De-platforming disinformation: conspiracy theories and their control}.
\newblock \bibinfo{journal}{\emph{Information, Communication \& Society}} \bibinfo{volume}{26}, \bibinfo{number}{6} (\bibinfo{year}{2023}), \bibinfo{pages}{1262--1280}.
\newblock


\bibitem[J{\"a}rvelin and Kek{\"a}l{\"a}inen(2002)]%
        {jarvelin2002cumulated}
\bibfield{author}{\bibinfo{person}{Kalervo J{\"a}rvelin} {and} \bibinfo{person}{Jaana Kek{\"a}l{\"a}inen}.} \bibinfo{year}{2002}\natexlab{}.
\newblock \showarticletitle{Cumulated gain-based evaluation of IR techniques}.
\newblock \bibinfo{journal}{\emph{ACM Transactions on Information Systems (TOIS)}} \bibinfo{volume}{20}, \bibinfo{number}{4} (\bibinfo{year}{2002}), \bibinfo{pages}{422--446}.
\newblock


\bibitem[Jiang et~al\mbox{.}(2021)]%
        {jiang2021mechanisms}
\bibfield{author}{\bibinfo{person}{Bohan Jiang}, \bibinfo{person}{Mansooreh Karami}, \bibinfo{person}{Lu Cheng}, \bibinfo{person}{Tyler Black}, {and} \bibinfo{person}{Huan Liu}.} \bibinfo{year}{2021}\natexlab{}.
\newblock \showarticletitle{Mechanisms and attributes of echo chambers in social media}.
\newblock \bibinfo{journal}{\emph{arXiv preprint arXiv:2106.05401}} (\bibinfo{year}{2021}).
\newblock


\bibitem[Jiang et~al\mbox{.}(2024)]%
        {jiang2024disinformation}
\bibfield{author}{\bibinfo{person}{Bohan Jiang}, \bibinfo{person}{Zhen Tan}, \bibinfo{person}{Ayushi Nirmal}, {and} \bibinfo{person}{Huan Liu}.} \bibinfo{year}{2024}\natexlab{}.
\newblock \showarticletitle{Disinformation detection: An evolving challenge in the age of llms}. In \bibinfo{booktitle}{\emph{Proceedings of the 2024 SIAM International Conference on Data Mining (SDM)}}. SIAM, \bibinfo{pages}{427--435}.
\newblock


\bibitem[Jolley and Paterson(2020)]%
        {jolley2020pylons}
\bibfield{author}{\bibinfo{person}{Daniel Jolley} {and} \bibinfo{person}{Jenny~L Paterson}.} \bibinfo{year}{2020}\natexlab{}.
\newblock \showarticletitle{Pylons ablaze: Examining the role of 5G COVID-19 conspiracy beliefs and support for violence}.
\newblock \bibinfo{journal}{\emph{British journal of social psychology}} \bibinfo{volume}{59}, \bibinfo{number}{3} (\bibinfo{year}{2020}), \bibinfo{pages}{628--640}.
\newblock


\bibitem[Kocmi and Federmann(2023)]%
        {kocmi2023large}
\bibfield{author}{\bibinfo{person}{Tom Kocmi} {and} \bibinfo{person}{Christian Federmann}.} \bibinfo{year}{2023}\natexlab{}.
\newblock \showarticletitle{Large Language Models Are State-of-the-Art Evaluators of Translation Quality}. In \bibinfo{booktitle}{\emph{Proceedings of the 24th Annual Conference of the European Association for Machine Translation}}. \bibinfo{pages}{193--203}.
\newblock


\bibitem[Kojima et~al\mbox{.}(2022)]%
        {kojima2022large}
\bibfield{author}{\bibinfo{person}{Takeshi Kojima}, \bibinfo{person}{Shixiang~Shane Gu}, \bibinfo{person}{Machel Reid}, \bibinfo{person}{Yutaka Matsuo}, {and} \bibinfo{person}{Yusuke Iwasawa}.} \bibinfo{year}{2022}\natexlab{}.
\newblock \showarticletitle{Large language models are zero-shot reasoners}.
\newblock \bibinfo{journal}{\emph{Advances in neural information processing systems}}  \bibinfo{volume}{35} (\bibinfo{year}{2022}), \bibinfo{pages}{22199--22213}.
\newblock


\bibitem[Lazer et~al\mbox{.}(2018)]%
        {lazer2018science}
\bibfield{author}{\bibinfo{person}{David~MJ Lazer}, \bibinfo{person}{Matthew~A Baum}, \bibinfo{person}{Yochai Benkler}, \bibinfo{person}{Adam~J Berinsky}, \bibinfo{person}{Kelly~M Greenhill}, \bibinfo{person}{Filippo Menczer}, \bibinfo{person}{Miriam~J Metzger}, \bibinfo{person}{Brendan Nyhan}, \bibinfo{person}{Gordon Pennycook}, \bibinfo{person}{David Rothschild}, {et~al\mbox{.}}} \bibinfo{year}{2018}\natexlab{}.
\newblock \showarticletitle{The science of fake news}.
\newblock \bibinfo{journal}{\emph{Science}} \bibinfo{volume}{359}, \bibinfo{number}{6380} (\bibinfo{year}{2018}), \bibinfo{pages}{1094--1096}.
\newblock


\bibitem[Li et~al\mbox{.}(2024a)]%
        {li2024generation}
\bibfield{author}{\bibinfo{person}{Dawei Li}, \bibinfo{person}{Bohan Jiang}, \bibinfo{person}{Liangjie Huang}, \bibinfo{person}{Alimohammad Beigi}, \bibinfo{person}{Chengshuai Zhao}, \bibinfo{person}{Zhen Tan}, \bibinfo{person}{Amrita Bhattacharjee}, \bibinfo{person}{Yuxuan Jiang}, \bibinfo{person}{Canyu Chen}, \bibinfo{person}{Tianhao Wu}, {et~al\mbox{.}}} \bibinfo{year}{2024}\natexlab{a}.
\newblock \showarticletitle{From Generation to Judgment: Opportunities and Challenges of LLM-as-a-judge}.
\newblock \bibinfo{journal}{\emph{arXiv preprint arXiv:2411.16594}} (\bibinfo{year}{2024}).
\newblock


\bibitem[Li et~al\mbox{.}(2024b)]%
        {li2024dalk}
\bibfield{author}{\bibinfo{person}{Dawei Li}, \bibinfo{person}{Shu Yang}, \bibinfo{person}{Zhen Tan}, \bibinfo{person}{Jae~Young Baik}, \bibinfo{person}{Sukwon Yun}, \bibinfo{person}{Joseph Lee}, \bibinfo{person}{Aaron Chacko}, \bibinfo{person}{Bojian Hou}, \bibinfo{person}{Duy Duong-Tran}, \bibinfo{person}{Ying Ding}, {et~al\mbox{.}}} \bibinfo{year}{2024}\natexlab{b}.
\newblock \showarticletitle{DALK: Dynamic Co-Augmentation of LLMs and KG to answer Alzheimer's Disease Questions with Scientific Literature}.
\newblock \bibinfo{journal}{\emph{arXiv preprint arXiv:2405.04819}} (\bibinfo{year}{2024}).
\newblock


\bibitem[Liang et~al\mbox{.}(2024)]%
        {liang2024can}
\bibfield{author}{\bibinfo{person}{Weixin Liang}, \bibinfo{person}{Yuhui Zhang}, \bibinfo{person}{Hancheng Cao}, \bibinfo{person}{Binglu Wang}, \bibinfo{person}{Daisy~Yi Ding}, \bibinfo{person}{Xinyu Yang}, \bibinfo{person}{Kailas Vodrahalli}, \bibinfo{person}{Siyu He}, \bibinfo{person}{Daniel~Scott Smith}, \bibinfo{person}{Yian Yin}, {et~al\mbox{.}}} \bibinfo{year}{2024}\natexlab{}.
\newblock \showarticletitle{Can large language models provide useful feedback on research papers? A large-scale empirical analysis}.
\newblock \bibinfo{journal}{\emph{NEJM AI}} \bibinfo{volume}{1}, \bibinfo{number}{8} (\bibinfo{year}{2024}), \bibinfo{pages}{AIoa2400196}.
\newblock


\bibitem[Liaw et~al\mbox{.}(2023)]%
        {liaw2023younicon}
\bibfield{author}{\bibinfo{person}{Shao~Yi Liaw}, \bibinfo{person}{Fan Huang}, \bibinfo{person}{Fabricio Benevenuto}, \bibinfo{person}{Haewoon Kwak}, {and} \bibinfo{person}{Jisun An}.} \bibinfo{year}{2023}\natexlab{}.
\newblock \showarticletitle{YouNICon: YouTube’s CommuNIty of Conspiracy Videos}. In \bibinfo{booktitle}{\emph{Proceedings of the International AAAI Conference on Web and Social Media}}, Vol.~\bibinfo{volume}{17}. \bibinfo{pages}{1102--1111}.
\newblock


\bibitem[Lin(2004)]%
        {lin2004rouge}
\bibfield{author}{\bibinfo{person}{Chin-Yew Lin}.} \bibinfo{year}{2004}\natexlab{}.
\newblock \showarticletitle{{ROUGE}: A Package for Automatic Evaluation of Summaries}. In \bibinfo{booktitle}{\emph{Text Summarization Branches Out}}. \bibinfo{publisher}{Association for Computational Linguistics}, \bibinfo{address}{Barcelona, Spain}, \bibinfo{pages}{74--81}.
\newblock
\urldef\tempurl%
\url{https://aclanthology.org/W04-1013}
\showURL{%
\tempurl}


\bibitem[Liu and Shah(2023)]%
        {liu2023reviewergpt}
\bibfield{author}{\bibinfo{person}{Ryan Liu} {and} \bibinfo{person}{Nihar~B Shah}.} \bibinfo{year}{2023}\natexlab{}.
\newblock \showarticletitle{Reviewergpt? an exploratory study on using large language models for paper reviewing}.
\newblock \bibinfo{journal}{\emph{arXiv preprint arXiv:2306.00622}} (\bibinfo{year}{2023}).
\newblock


\bibitem[Lyu et~al\mbox{.}(2023)]%
        {lyu2023gpt}
\bibfield{author}{\bibinfo{person}{Hanjia Lyu}, \bibinfo{person}{Jinfa Huang}, \bibinfo{person}{Daoan Zhang}, \bibinfo{person}{Yongsheng Yu}, \bibinfo{person}{Xinyi Mou}, \bibinfo{person}{Jinsheng Pan}, \bibinfo{person}{Zhengyuan Yang}, \bibinfo{person}{Zhongyu Wei}, {and} \bibinfo{person}{Jiebo Luo}.} \bibinfo{year}{2023}\natexlab{}.
\newblock \showarticletitle{Gpt-4v (ision) as a social media analysis engine}.
\newblock \bibinfo{journal}{\emph{arXiv preprint arXiv:2311.07547}} (\bibinfo{year}{2023}).
\newblock


\bibitem[McAleese et~al\mbox{.}(2024)]%
        {mcaleese2024llm}
\bibfield{author}{\bibinfo{person}{Nat McAleese}, \bibinfo{person}{Rai~Michael Pokorny}, \bibinfo{person}{Juan Felipe~Ceron Uribe}, \bibinfo{person}{Evgenia Nitishinskaya}, \bibinfo{person}{Maja Trebacz}, {and} \bibinfo{person}{Jan Leike}.} \bibinfo{year}{2024}\natexlab{}.
\newblock \showarticletitle{Llm critics help catch llm bugs}.
\newblock \bibinfo{journal}{\emph{arXiv preprint arXiv:2407.00215}} (\bibinfo{year}{2024}).
\newblock


\bibitem[{Media Bias/Fact Check}(2024)]%
        {mediabias_yougov}
\bibfield{author}{\bibinfo{person}{{Media Bias/Fact Check}}.} \bibinfo{year}{2024}\natexlab{}.
\newblock \bibinfo{booktitle}{\emph{YouGov Polling US Bias and Credibility}}.
\newblock
\urldef\tempurl%
\url{https://mediabiasfactcheck.com/yougov-polling-us-bias-and-credibility/}
\showURL{%
\tempurl}
\newblock
\shownote{Accessed: 2024-12-05}.


\bibitem[Monti et~al\mbox{.}(2023)]%
        {monti2023online}
\bibfield{author}{\bibinfo{person}{Corrado Monti}, \bibinfo{person}{Matteo Cinelli}, \bibinfo{person}{Carlo Valensise}, \bibinfo{person}{Walter Quattrociocchi}, {and} \bibinfo{person}{Michele Starnini}.} \bibinfo{year}{2023}\natexlab{}.
\newblock \showarticletitle{Online conspiracy communities are more resilient to deplatforming}.
\newblock \bibinfo{journal}{\emph{PNAS nexus}} \bibinfo{volume}{2}, \bibinfo{number}{10} (\bibinfo{year}{2023}), \bibinfo{pages}{pgad324}.
\newblock


\bibitem[OpenAI(2024)]%
        {openai2024o1preview}
\bibfield{author}{\bibinfo{person}{OpenAI}.} \bibinfo{year}{2024}\natexlab{}.
\newblock \bibinfo{title}{Introducing OpenAI O1 Preview}.
\newblock
\newblock
\urldef\tempurl%
\url{https://openai.com/index/introducing-openai-o1-preview/}
\showURL{%
\tempurl}
\newblock
\shownote{Accessed: 2024-11-01}.


\bibitem[Papineni et~al\mbox{.}(2002)]%
        {papineni2002bleu}
\bibfield{author}{\bibinfo{person}{Kishore Papineni}, \bibinfo{person}{Salim Roukos}, \bibinfo{person}{Todd Ward}, {and} \bibinfo{person}{Wei-Jing Zhu}.} \bibinfo{year}{2002}\natexlab{}.
\newblock \showarticletitle{{B}leu: a Method for Automatic Evaluation of Machine Translation}. In \bibinfo{booktitle}{\emph{Proceedings of the 40th Annual Meeting of the Association for Computational Linguistics}}, \bibfield{editor}{\bibinfo{person}{Pierre Isabelle}, \bibinfo{person}{Eugene Charniak}, {and} \bibinfo{person}{Dekang Lin}} (Eds.). \bibinfo{publisher}{Association for Computational Linguistics}, \bibinfo{address}{Philadelphia, Pennsylvania, USA}, \bibinfo{pages}{311--318}.
\newblock
\urldef\tempurl%
\url{https://doi.org/10.3115/1073083.1073135}
\showDOI{\tempurl}


\bibitem[Park et~al\mbox{.}(2023)]%
        {park2023generative}
\bibfield{author}{\bibinfo{person}{Joon~Sung Park}, \bibinfo{person}{Joseph O'Brien}, \bibinfo{person}{Carrie~Jun Cai}, \bibinfo{person}{Meredith~Ringel Morris}, \bibinfo{person}{Percy Liang}, {and} \bibinfo{person}{Michael~S Bernstein}.} \bibinfo{year}{2023}\natexlab{}.
\newblock \showarticletitle{Generative agents: Interactive simulacra of human behavior}. In \bibinfo{booktitle}{\emph{Proceedings of the 36th annual acm symposium on user interface software and technology}}. \bibinfo{pages}{1--22}.
\newblock


\bibitem[Pertwee et~al\mbox{.}(2022)]%
        {pertwee2022epidemic}
\bibfield{author}{\bibinfo{person}{Ed Pertwee}, \bibinfo{person}{Clarissa Simas}, {and} \bibinfo{person}{Heidi~J Larson}.} \bibinfo{year}{2022}\natexlab{}.
\newblock \showarticletitle{An epidemic of uncertainty: rumors, conspiracy theories and vaccine hesitancy}.
\newblock \bibinfo{journal}{\emph{Nature medicine}} \bibinfo{volume}{28}, \bibinfo{number}{3} (\bibinfo{year}{2022}), \bibinfo{pages}{456--459}.
\newblock


\bibitem[Romer and Jamieson(2020)]%
        {romer2020conspiracy}
\bibfield{author}{\bibinfo{person}{Daniel Romer} {and} \bibinfo{person}{Kathleen~Hall Jamieson}.} \bibinfo{year}{2020}\natexlab{}.
\newblock \showarticletitle{Conspiracy theories as barriers to controlling the spread of COVID-19 in the US}.
\newblock \bibinfo{journal}{\emph{Social science \& medicine}}  \bibinfo{volume}{263} (\bibinfo{year}{2020}), \bibinfo{pages}{113356}.
\newblock


\bibitem[Romer and Jamieson(2021)]%
        {romer2021patterns}
\bibfield{author}{\bibinfo{person}{Daniel Romer} {and} \bibinfo{person}{Kathleen~Hall Jamieson}.} \bibinfo{year}{2021}\natexlab{}.
\newblock \showarticletitle{Patterns of media use, strength of belief in COVID-19 conspiracy theories, and the prevention of COVID-19 from March to July 2020 in the United States: survey study}.
\newblock \bibinfo{journal}{\emph{Journal of medical Internet research}} \bibinfo{volume}{23}, \bibinfo{number}{4} (\bibinfo{year}{2021}), \bibinfo{pages}{e25215}.
\newblock


\bibitem[Sen(1968)]%
        {sen1968estimates}
\bibfield{author}{\bibinfo{person}{Pranab~Kumar Sen}.} \bibinfo{year}{1968}\natexlab{}.
\newblock \showarticletitle{Estimates of the regression coefficient based on Kendall's tau}.
\newblock \bibinfo{journal}{\emph{Journal of the American statistical association}} \bibinfo{volume}{63}, \bibinfo{number}{324} (\bibinfo{year}{1968}), \bibinfo{pages}{1379--1389}.
\newblock


\bibitem[Shahsavari et~al\mbox{.}(2020)]%
        {shahsavari2020conspiracy}
\bibfield{author}{\bibinfo{person}{Shadi Shahsavari}, \bibinfo{person}{Pavan Holur}, \bibinfo{person}{Tianyi Wang}, \bibinfo{person}{Timothy~R Tangherlini}, {and} \bibinfo{person}{Vwani Roychowdhury}.} \bibinfo{year}{2020}\natexlab{}.
\newblock \showarticletitle{Conspiracy in the time of corona: automatic detection of emerging COVID-19 conspiracy theories in social media and the news}.
\newblock \bibinfo{journal}{\emph{Journal of computational social science}} \bibinfo{volume}{3}, \bibinfo{number}{2} (\bibinfo{year}{2020}), \bibinfo{pages}{279--317}.
\newblock


\bibitem[Spearman(1961)]%
        {spearman1961proof}
\bibfield{author}{\bibinfo{person}{Charles Spearman}.} \bibinfo{year}{1961}\natexlab{}.
\newblock \showarticletitle{The proof and measurement of association between two things.}
\newblock  (\bibinfo{year}{1961}).
\newblock


\bibitem[Sunstein and Vermeule(2009)]%
        {sunstein2009conspiracy}
\bibfield{author}{\bibinfo{person}{Cass~R Sunstein} {and} \bibinfo{person}{Adrian Vermeule}.} \bibinfo{year}{2009}\natexlab{}.
\newblock \showarticletitle{Conspiracy theories: causes and cures.}
\newblock \bibinfo{journal}{\emph{Journal of political philosophy}} \bibinfo{volume}{17}, \bibinfo{number}{2} (\bibinfo{year}{2009}).
\newblock


\bibitem[Tong et~al\mbox{.}(2024)]%
        {tong2024can}
\bibfield{author}{\bibinfo{person}{Yongqi Tong}, \bibinfo{person}{Dawei Li}, \bibinfo{person}{Sizhe Wang}, \bibinfo{person}{Yujia Wang}, \bibinfo{person}{Fei Teng}, {and} \bibinfo{person}{Jingbo Shang}.} \bibinfo{year}{2024}\natexlab{}.
\newblock \showarticletitle{Can LLMs Learn from Previous Mistakes? Investigating LLMs' Errors to Boost for Reasoning}.
\newblock \bibinfo{journal}{\emph{arXiv preprint arXiv:2403.20046}} (\bibinfo{year}{2024}).
\newblock


\bibitem[T{\"o}rnberg et~al\mbox{.}(2023)]%
        {tornberg2023simulating}
\bibfield{author}{\bibinfo{person}{Petter T{\"o}rnberg}, \bibinfo{person}{Diliara Valeeva}, \bibinfo{person}{Justus Uitermark}, {and} \bibinfo{person}{Christopher Bail}.} \bibinfo{year}{2023}\natexlab{}.
\newblock \showarticletitle{Simulating social media using large language models to evaluate alternative news feed algorithms}.
\newblock \bibinfo{journal}{\emph{arXiv preprint arXiv:2310.05984}} (\bibinfo{year}{2023}).
\newblock


\bibitem[Touvron et~al\mbox{.}(2023a)]%
        {touvron2023llama}
\bibfield{author}{\bibinfo{person}{Hugo Touvron}, \bibinfo{person}{Thibaut Lavril}, \bibinfo{person}{Gautier Izacard}, \bibinfo{person}{Xavier Martinet}, \bibinfo{person}{Marie-Anne Lachaux}, \bibinfo{person}{Timoth{\'e}e Lacroix}, \bibinfo{person}{Baptiste Rozi{\`e}re}, \bibinfo{person}{Naman Goyal}, \bibinfo{person}{Eric Hambro}, \bibinfo{person}{Faisal Azhar}, {et~al\mbox{.}}} \bibinfo{year}{2023}\natexlab{a}.
\newblock \showarticletitle{Llama: Open and efficient foundation language models}.
\newblock \bibinfo{journal}{\emph{arXiv preprint arXiv:2302.13971}} (\bibinfo{year}{2023}).
\newblock


\bibitem[Touvron et~al\mbox{.}(2023b)]%
        {touvron2023llama2}
\bibfield{author}{\bibinfo{person}{Hugo Touvron}, \bibinfo{person}{Louis Martin}, \bibinfo{person}{Kevin Stone}, \bibinfo{person}{Peter Albert}, \bibinfo{person}{Amjad Almahairi}, \bibinfo{person}{Yasmine Babaei}, \bibinfo{person}{Nikolay Bashlykov}, \bibinfo{person}{Soumya Batra}, \bibinfo{person}{Prajjwal Bhargava}, \bibinfo{person}{Shruti Bhosale}, {et~al\mbox{.}}} \bibinfo{year}{2023}\natexlab{b}.
\newblock \showarticletitle{Llama 2: Open foundation and fine-tuned chat models}.
\newblock \bibinfo{journal}{\emph{arXiv preprint arXiv:2307.09288}} (\bibinfo{year}{2023}).
\newblock


\bibitem[Wang et~al\mbox{.}(2024b)]%
        {wang2024bpo}
\bibfield{author}{\bibinfo{person}{Sizhe Wang}, \bibinfo{person}{Yongqi Tong}, \bibinfo{person}{Hengyuan Zhang}, \bibinfo{person}{Dawei Li}, \bibinfo{person}{Xin Zhang}, {and} \bibinfo{person}{Tianlong Chen}.} \bibinfo{year}{2024}\natexlab{b}.
\newblock \showarticletitle{Bpo: Towards balanced preference optimization between knowledge breadth and depth in alignment}.
\newblock \bibinfo{journal}{\emph{arXiv preprint arXiv:2411.10914}} (\bibinfo{year}{2024}).
\newblock


\bibitem[Wang et~al\mbox{.}(2024a)]%
        {wang2024llm}
\bibfield{author}{\bibinfo{person}{Zilong Wang}, \bibinfo{person}{Xufang Luo}, \bibinfo{person}{Xinyang Jiang}, \bibinfo{person}{Dongsheng Li}, {and} \bibinfo{person}{Lili Qiu}.} \bibinfo{year}{2024}\natexlab{a}.
\newblock \showarticletitle{LLM-RadJudge: Achieving Radiologist-Level Evaluation for X-Ray Report Generation}.
\newblock \bibinfo{journal}{\emph{arXiv preprint arXiv:2404.00998}} (\bibinfo{year}{2024}).
\newblock


\bibitem[Yang et~al\mbox{.}(2024a)]%
        {yang2024qwen2}
\bibfield{author}{\bibinfo{person}{An Yang}, \bibinfo{person}{Baosong Yang}, \bibinfo{person}{Binyuan Hui}, \bibinfo{person}{Bo Zheng}, \bibinfo{person}{Bowen Yu}, \bibinfo{person}{Chang Zhou}, \bibinfo{person}{Chengpeng Li}, \bibinfo{person}{Chengyuan Li}, \bibinfo{person}{Dayiheng Liu}, \bibinfo{person}{Fei Huang}, {et~al\mbox{.}}} \bibinfo{year}{2024}\natexlab{a}.
\newblock \showarticletitle{Qwen2 technical report}.
\newblock \bibinfo{journal}{\emph{arXiv preprint arXiv:2407.10671}} (\bibinfo{year}{2024}).
\newblock


\bibitem[Yang et~al\mbox{.}(2024b)]%
        {yang2024oasis}
\bibfield{author}{\bibinfo{person}{Ziyi Yang}, \bibinfo{person}{Zaibin Zhang}, \bibinfo{person}{Zirui Zheng}, \bibinfo{person}{Yuxian Jiang}, \bibinfo{person}{Ziyue Gan}, \bibinfo{person}{Zhiyu Wang}, \bibinfo{person}{Zijian Ling}, \bibinfo{person}{Jinsong Chen}, \bibinfo{person}{Martz Ma}, \bibinfo{person}{Bowen Dong}, {et~al\mbox{.}}} \bibinfo{year}{2024}\natexlab{b}.
\newblock \showarticletitle{OASIS: Open Agents Social Interaction Simulations on One Million Agents}.
\newblock \bibinfo{journal}{\emph{arXiv preprint arXiv:2411.11581}} (\bibinfo{year}{2024}).
\newblock


\bibitem[Ye et~al\mbox{.}(2024)]%
        {ye2024we}
\bibfield{author}{\bibinfo{person}{Rui Ye}, \bibinfo{person}{Xianghe Pang}, \bibinfo{person}{Jingyi Chai}, \bibinfo{person}{Jiaao Chen}, \bibinfo{person}{Zhenfei Yin}, \bibinfo{person}{Zhen Xiang}, \bibinfo{person}{Xiaowen Dong}, \bibinfo{person}{Jing Shao}, {and} \bibinfo{person}{Siheng Chen}.} \bibinfo{year}{2024}\natexlab{}.
\newblock \showarticletitle{Are We There Yet? Revealing the Risks of Utilizing Large Language Models in Scholarly Peer Review}.
\newblock \bibinfo{journal}{\emph{arXiv preprint arXiv:2412.01708}} (\bibinfo{year}{2024}).
\newblock


\bibitem[YouGov(2023)]%
        {yougov2023conspiracy}
\bibfield{author}{\bibinfo{person}{YouGov}.} \bibinfo{year}{2023}\natexlab{}.
\newblock \bibinfo{booktitle}{\emph{YouGov Survey: Conspiracy Theories}}.
\newblock
\urldef\tempurl%
\url{https://d3nkl3psvxxpe9.cloudfront.net/documents/Conspiracy_Theories_poll_results.pdf}
\showURL{%
Retrieved Sep 17, 2024 from \tempurl}


\bibitem[Yu et~al\mbox{.}(2024)]%
        {yu2024popalm}
\bibfield{author}{\bibinfo{person}{Erxin Yu}, \bibinfo{person}{Jing Li}, {and} \bibinfo{person}{Chunpu Xu}.} \bibinfo{year}{2024}\natexlab{}.
\newblock \showarticletitle{PopALM: Popularity-Aligned Language Models for Social Media Trendy Response Prediction}.
\newblock \bibinfo{journal}{\emph{arXiv preprint arXiv:2402.18950}} (\bibinfo{year}{2024}).
\newblock


\bibitem[Zhao et~al\mbox{.}(2024)]%
        {zhao2024codejudge}
\bibfield{author}{\bibinfo{person}{Yuwei Zhao}, \bibinfo{person}{Ziyang Luo}, \bibinfo{person}{Yuchen Tian}, \bibinfo{person}{Hongzhan Lin}, \bibinfo{person}{Weixiang Yan}, \bibinfo{person}{Annan Li}, {and} \bibinfo{person}{Jing Ma}.} \bibinfo{year}{2024}\natexlab{}.
\newblock \showarticletitle{CodeJudge-Eval: Can Large Language Models be Good Judges in Code Understanding?}
\newblock \bibinfo{journal}{\emph{arXiv preprint arXiv:2408.10718}} (\bibinfo{year}{2024}).
\newblock


\bibitem[Zheng et~al\mbox{.}(2023)]%
        {zheng2023judging}
\bibfield{author}{\bibinfo{person}{Lianmin Zheng}, \bibinfo{person}{Wei-Lin Chiang}, \bibinfo{person}{Ying Sheng}, \bibinfo{person}{Siyuan Zhuang}, \bibinfo{person}{Zhanghao Wu}, \bibinfo{person}{Yonghao Zhuang}, \bibinfo{person}{Zi Lin}, \bibinfo{person}{Zhuohan Li}, \bibinfo{person}{Dacheng Li}, \bibinfo{person}{Eric Xing}, {et~al\mbox{.}}} \bibinfo{year}{2023}\natexlab{}.
\newblock \showarticletitle{Judging llm-as-a-judge with mt-bench and chatbot arena}.
\newblock \bibinfo{journal}{\emph{Advances in Neural Information Processing Systems}}  \bibinfo{volume}{36} (\bibinfo{year}{2023}), \bibinfo{pages}{46595--46623}.
\newblock


\bibitem[Zhou et~al\mbox{.}(2024)]%
        {zhou2024llm}
\bibfield{author}{\bibinfo{person}{Ruiyang Zhou}, \bibinfo{person}{Lu Chen}, {and} \bibinfo{person}{Kai Yu}.} \bibinfo{year}{2024}\natexlab{}.
\newblock \showarticletitle{Is LLM a Reliable Reviewer? A Comprehensive Evaluation of LLM on Automatic Paper Reviewing Tasks}. In \bibinfo{booktitle}{\emph{Proceedings of the 2024 Joint International Conference on Computational Linguistics, Language Resources and Evaluation (LREC-COLING 2024)}}. \bibinfo{pages}{9340--9351}.
\newblock


\bibitem[Zhou et~al\mbox{.}(2023)]%
        {zhou2023sotopia}
\bibfield{author}{\bibinfo{person}{Xuhui Zhou}, \bibinfo{person}{Hao Zhu}, \bibinfo{person}{Leena Mathur}, \bibinfo{person}{Ruohong Zhang}, \bibinfo{person}{Haofei Yu}, \bibinfo{person}{Zhengyang Qi}, \bibinfo{person}{Louis-Philippe Morency}, \bibinfo{person}{Yonatan Bisk}, \bibinfo{person}{Daniel Fried}, \bibinfo{person}{Graham Neubig}, {et~al\mbox{.}}} \bibinfo{year}{2023}\natexlab{}.
\newblock \showarticletitle{Sotopia: Interactive evaluation for social intelligence in language agents}.
\newblock \bibinfo{journal}{\emph{arXiv preprint arXiv:2310.11667}} (\bibinfo{year}{2023}).
\newblock


\bibitem[Ziems et~al\mbox{.}(2024)]%
        {ziems2024can}
\bibfield{author}{\bibinfo{person}{Caleb Ziems}, \bibinfo{person}{William Held}, \bibinfo{person}{Omar Shaikh}, \bibinfo{person}{Jiaao Chen}, \bibinfo{person}{Zhehao Zhang}, {and} \bibinfo{person}{Diyi Yang}.} \bibinfo{year}{2024}\natexlab{}.
\newblock \showarticletitle{Can large language models transform computational social science?}
\newblock \bibinfo{journal}{\emph{Computational Linguistics}} \bibinfo{volume}{50}, \bibinfo{number}{1} (\bibinfo{year}{2024}), \bibinfo{pages}{237--291}.
\newblock


\end{thebibliography}

\end{document}